\documentclass[10pt,twocolumn,letterpaper]{article}

\usepackage{cvpr}
\usepackage{times}
\usepackage{epsfig}
\usepackage{graphicx}
\usepackage{amsmath,amssymb}
\usepackage{bbm}
\usepackage{multirow}
\usepackage{caption}
\usepackage{algorithm}
\usepackage{algpseudocode}
\usepackage{dsfont}
\usepackage{cite}

\usepackage{bm}
\usepackage{booktabs}
\usepackage{color}

\usepackage[breaklinks=true,bookmarks=false]{hyperref}

\cvprfinalcopy 

\def\cvprPaperID{****} 

\begin{document}

\title{Skeleton-Based Online Action Prediction Using Scale Selection Network}

\author{Jun~Liu$^{1}$\\{\tt\small jliu029@ntu.edu.sg}
  \and Amir Shahroudy$^{2}$\\{\tt\small amirsh@chalmers.se}
  \and Gang~Wang$^{3}$\\{\tt\small gangwang6@gmail.com}
	\and Ling-Yu~Duan$^{4}$\\{\tt\small lingyu@pku.edu.cn}
  \and Alex~C.~Kot$^{1}$\\{\tt\small eackot@ntu.edu.sg}
	\and $^1$ School of Electrical and Electronic Engineering, Nanyang Technological University, Singapore
   \\$^2$ Chalmers University of Technology, Sweden
   \\$^3$ Alibaba Group, China \hspace{20pt} 
   \\$^4$ Peking University, China
}

\maketitle

\begin{abstract}
   Action prediction is to recognize the class label of an ongoing activity when only a part of it is observed.
   In this paper, we focus on online action prediction in streaming 3D skeleton sequences.
   A dilated convolutional network is introduced to model the motion dynamics in temporal dimension via a sliding window over the temporal axis.
   Since there are significant temporal scale variations in the observed part of the ongoing action at different time steps,
   a novel window scale selection method is proposed to make our network focus on the performed part of the ongoing action and try to suppress the possible incoming interference from the previous actions
   at each step.
   An activation sharing scheme is also proposed to handle the overlapping computations among the adjacent time steps,
   which enables our framework to run more efficiently.
   Moreover, to enhance the performance of our framework for action prediction with the skeletal input data,
   a hierarchy of dilated tree convolutions are also designed to learn the multi-level structured semantic representations over the skeleton joints at each frame.
   Our proposed approach is evaluated on four challenging datasets.
   The extensive experiments demonstrate the effectiveness of our method for skeleton-based online action prediction.  
\end{abstract}

\section{Introduction}
\label{sec:introduction}
\begin{figure}[t]
	\centerline{\includegraphics[scale=0.3]{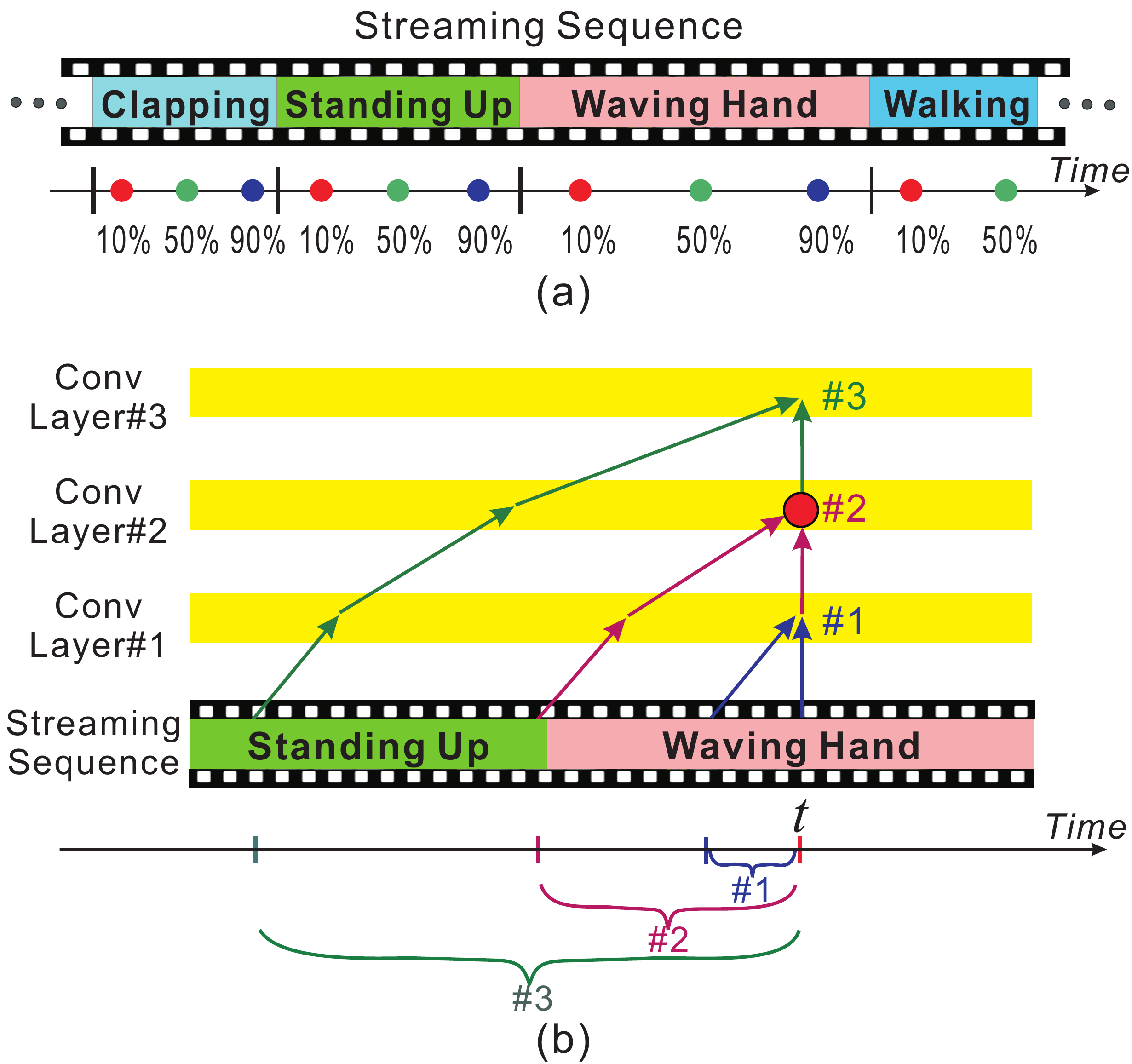}}
	\caption{ Figure (a) illustrates an untrimmed streaming sequence that contains multiple action instances.
            We need to recognize the current ongoing action at each time step when only a part (\eg, $10\%$) of it is performed.
            Figure (b) depicts our SSNet approach for online action prediction.
            At time $t$, only a part of the action \emph{waving hand} is observed.
            Our SSNet selects the convolutional layer \#2 rather than \#3 for prediction,
            as the perception window of \#2 mainly covers the performed part of current action,
            while \#3 involves too many frames from the previous action which can interfere the prediction at time step $t$.}
	\label{fig:coverfig}
\end{figure}

In action prediction (early action recognition), the goal is to predict the class label of an ongoing action from an observed part of it over temporal axis so far.
Predicting actions before they get completely performed is a subset of a broader research domain on human activity analysis.
It has attracted a lot of research attention due to its wide range of applications in
security surveillance, human-machine interaction, patient monitoring, etc \cite{hu2016real,cao2013recognize}.
Most of the existing works \cite{hu2016real,ke2016human,kongdeep2017} focus on action prediction in well-segmented videos, for which each video contains exactly one action instance.
However, in more practical scenarios, such as online human-machine interaction systems,
plenty of unsegmented action instances are contained in a streaming sequence.
In this paper, we address this challenging task: ``online action prediction in untrimmed video'',
\ie, we aim to recognize the current ongoing action from the observed part of it at each time step of the data stream,
which can include multiple actions, as illustrated in \figurename{~\ref{fig:coverfig}(a)}.

The biological studies \cite{johansson1973visual} demonstrate that skeleton data is informative enough for representing human behavior,
even without appearance information \cite{zhang2017view}.
Human activities are naturally performed in 3D space, thus 3D skeleton data is suitable for representing human actions \cite{mawalking2017}.
The 3D skeleton information can be easily and effectively acquired in real-time with the low-cost depth sensors \cite{han2013enhanced}, such as Microsoft Kinect and Asus Xtion. 
As a result, activity analysis with 3D skeleton data becomes a popular domain of research
\cite{han2017review,presti20163d,veeriah2015differential,liu2017PAMI,liu2017fusing,zhang2017geometric,liu2018TIP,rahmani2017learning,rahmani2014real}
thanks to its succinctness, high level representation,
and robustness against variations in viewpoints, illumination, clothing textures, and background clutter \cite{du2015hierarchical,liu2016eccv,hu2016real}.

We investigate real-time action prediction with the continuous 3D skeleton data in this paper.
To predict the class label of the current ongoing action at each time step,
we adopt a sliding window over the temporal axis of the input streams of skeleton sequences,
and the frames under the window are used as input to perform action prediction.

The sliding window design has been widely employed for a series of vision related tasks,
such as object recognition \cite{mutch2006multiclass},
pedestrian detection \cite{enzweiler2009monocular},
activity detection \cite{oneata2014lear,siva2011weakly,zanfir2013moving,hoai2014max}, etc.
Most of these works utilize one fixed scale, or combine multi-scale multi-pass scans at each sliding position.
However, in our online action prediction task, we need to predict the ongoing action at each observation ratio,
while there are significant temporal scale variations in the observed part of the ongoing action. 
This makes it difficult to determine the scale of the sliding window.

The untrimmed streaming sequence may contain multiple action instances, as shown in \figurename{~\ref{fig:coverfig}}(a).
The order of the actions can be arbitrary,
and the duration of different instances is often not the same.
Moreover, the observed (per whole) ratio of the ongoing action changes over time,
which makes it even more challenging to obtain a proper temporal window scale for online prediction.
For instance, at an early temporal stage,
it is beneficial to use a relatively smaller temporal window,
because the larger window sizes may include frames from the previous action instances which can mislead the recognition of the current instance.
Conversely, if a large part of the current action has already been observed,
it is beneficial to use a larger window size to cover more of its performed parts in order to achieve a reliable prediction.

To tackle the aforementioned challenges, in this paper, a novel Scale Selection Network (SSNet) is proposed for online action prediction.
Instead of using a fixed scale or multi-scale multi-pass scans at each time step,
we supervise our network to dynamically learn the proper temporal window scale at each step to cover the performed part of the current action instance.
In our approach,
the network predicts the ongoing action at each frame.
Beside predicting the class label,
it also regresses the temporal distance to the beginning of current action instance,
which indicates the performed part of the ongoing action.
Thus, at the next temporal step (next frame), we can utilize this value as the temporal window scale for action class prediction.

In our network, we apply convolutional analysis in temporal dimension to model the motion dynamics over the frames for skeleton-based action prediction.
A hierarchical architecture with dilated convolution filters is leveraged to learn a comprehensive representation over the frames within each perception window,
such that different layers in our SSNet correspond to different temporal scales,
as shown in \figurename{~\ref{fig:coverfig}(b)}.
Therefore, at each time step, our network selects the \emph{proper} convolutional layer which covers the most similar window scale regressed by its previous step.
Then the activations of this layer can be used for action prediction.
The proposed SSNet is designed to select the proper window in order to cover the performed part of the current action
and try to suppress the unrelated data from the previous ones.
Hence it produces reliable predictions at each step.
To the best of our knowledge,
this is the first convolutional model with explicit temporal scale selection as its fundamental capability for handling scale variations in online activity analysis.

In many existing approaches that utilize sliding window designs,
the computational efficiency is often relatively low due to the overlapping design and exhaustive multi-scale multi-round scans.
In our method, the action prediction is performed with a regressed scale at each step, which avoids multi-pass scans.
So the action prediction and scale selection are performed by a single convolutional network very efficiently.
Moreover, we introduce an activation sharing scheme to deal with the overlapping computations over different time steps,
which makes our SSNet run very fast for real-time online prediction.

In addition, to improve the performance of our network in handling the 3D skeleton data as input,
we also propose a hierarchy of dilated tree convolutions to learn the multi-level structured semantic representations over the skeleton joints at each frame for our action prediction network.

The main contributions of this paper are summarized as follows:

\begin{enumerate}
  \item
We study the new problem of real-time online action prediction in continuous 3D skeleton streams by leveraging convolutional analysis in temporal dimension.
  \item
Our proposed SSNet is capable of dealing with the scale variations of the observed portion of the ongoing action at different time steps.
We propose a scale selection scheme to let our network learn the proper temporal scale at each step,
such that the network can mainly focus on the performed part of the current action, and try to avoid the clutter from the previous actions data in the input online stream.
  \item
A hierarchy of dilated tree convolutions are also proposed
to learn multi-level structured representations for the input skeleton data and improve the performance of our SSNet for skeleton-based action prediction.
  \item
The proposed framework is very efficient for online action analysis thanks to the computation sharing over different time steps.
  \item
We perform action prediction with our SSNet which is end-to-end trainable, rather than using expensive multi-stage multi-network design at each step.
  \item
The proposed method achieves superior performance on four challenging datasets for 3D skeleton-based activity analysis.
\end{enumerate}

The remainder of this paper is organized as follows.
We review the related works in section \ref{sec:relatedwork}.
In section \ref{sec:method}, we introduce our proposed SSNet for skeleton-based online action prediction in detail.
We present the experimental results and comparisons in section \ref{sec:experiments}.
Finally, we conclude the paper in section \ref{sec:conclusion}.

\section{Related Work}
\label{sec:relatedwork}

\textbf{Skeleton-Based Action Recognition.}
With the advent of cheap and easy-to-use depth sensors, such as Kinect \cite{han2013enhanced},
3D skeleton-based human action recognition became a popular research domain \cite{aggarwal2014human,zhang2016rgb},
and a series of hand-crafted features \cite{eigenjointsJournal,actionletPAMI,vemulapalli2014liegroup,shahroudy2016multimodal,skeletalQuads,yun2012two,yu2014discriminative,hu2017jointly,HOJ3D,wang2016graph}
and deep learning-based approaches \cite{du2015hierarchical,ke2018computer,zhu2016co,Shahroudy_2016_CVPR,li2018spatio,kim2017interpretable,liu2016eccv,yan2018spatial,ke2017skeletonnet} have been proposed.

Most of the existing skeleton-based action recognition methods \cite{vemulapalli2014liegroup,Shahroudy_2016_CVPR,huang2017deep,wang2016action,du2015skeleton,ke2017new}
receive the fully observed segmented videos as input (each sample contains one full action instance), and derive a class label.
The proposed skeleton-based online action prediction method takes one step forward in dealing with numerous action instances occurring in the untrimmed sequences,
for which the current ongoing action can be only partly observed.
There are a limited number of skeleton-based action recognition methods \cite{bloom2012g3d} for untrimmed online sequences.
Different from these works, the proposed SSNet framework predicts the class label of the current ongoing action by utilizing its predicted observation ratio.

\textbf{Action Prediction.}
Predicting (recognizing) an action before it gets fully performed has attracted a lot of research attention recently \cite{kong2014discriminative,ryoo2011human,cao2013recognize,xu2015activity,li2014prediction,ke2016human,ke2017leveraging}.

Cao \etal \cite{cao2013recognize} formulated the prediction task as a posterior-maximization problem,
and applied sparse coding for action prediction.
Ryoo \etal \cite{ryoo2011human} represented each action as an integral histogram of spatio-temporal features.
They also developed a recognition methodology called dynamic bag-of-words (DBoW) for activity prediction.
Li \etal \cite{li2012modeling} designed a predictive accumulative function.
In their method, the human activities are represented as a temporal composition of constituent actionlets.
Kong \etal \cite{kong2014discriminative} proposed a discriminative multi-scale model for early action recognition.
Ke \etal \cite{ke2016human} extracted deep features in optical flow images for activity prediction.

Hu \etal \cite{hu2016real} explored to incorporate 3D skeleton information for real-time action prediction in the \emph{well-segmented} sequences,
\ie, each sequence includes only one action.
They introduced a soft regression strategy for action prediction.
An accumulative frame feature was also designed to make their method work efficiently.
However, their framework is not suitable for online action prediction in the \emph{untrimmed} continuous skeleton sequence
that contains multiple action instances.



\textbf{Action Analysis with Untrimmed Sequences.}
Beside the online action prediction task,
the problem of temporal action detection \cite{wei2013concurrent,yeung2016end,gao2017turn,oneata2014lear,lea2016temporal,dai2017temporal,sharaf2015real,gao2017red,wei2013concurrent,shou2017cdc,gao2017cascaded,wang2017untrimmednets,zhao2017temporal} also copes with untrimmed videos.
Several methods attempted online detection \cite{li2016online},
while most of the action detection approaches are developed for handling offline mode that conducts detection after observing the whole long sequence \cite{oneata2014lear,li2017skeleton,siva2011weakly}.

Our task is different from action detection,
as action detection mainly addresses accurate spatio-temporal segmentation,
while action prediction focuses more on predicting the class of the current ongoing action from its observed part so far,
even when only a small ratio of it is performed. 

Sliding window-based design \cite{zanfir2013moving,sharaf2015real,baek2017real,hoai2014max} and action proposals \cite{dai2017temporal} have been adopted for action detection.
Zanfir \etal \cite{zanfir2013moving} used a sliding window with one fixed scale (obtained by cross validation) for action detection.
Shou \etal \cite{shou2016temporal} adopted multi-scale windows for action detection via multi-stage networks.

Differently, in our online action prediction task,
determining the scale of the temporal window is challenging due to the scale variations of the observed part of the ongoing action. 
Also, rather than using one fixed scale \cite{zanfir2013moving} or multi-scale multi-round scans \cite{shou2016temporal,zhu2017efficient},
we propose a novel SSNet for online prediction, which is supervised to choose the proper window for prediction at each time step.
Moreover, the redundant computations are efficiently shared over different steps in our approach. 

This manuscript is the extension of our recent conference paper \cite{liu2018ssnet}.
The contributions of this work over \cite{liu2018ssnet} are as follows.
In \cite{liu2018ssnet}, the coordinates of the skeleton joints at each frame were simply concatenated to form a vector representing the current frame's pose.
Such a representation ignores the underlying semantics of spatial pose structures.
In this paper, we propose a hierarchy of dilated tree convolutions to process the input data
and learn more powerful multi-level structured semantic representations at each frame of the streaming skeleton sequence.
The newly proposed multi-level structured representation enhances the capability of our framework for action prediction in 3D skeleton streams.
In addition, we provide a more in-depth description of the proposed method and its implementation details.
Furthermore, we extensively evaluate the proposed action prediction framework on two more datasets,
including the large-scale ChaLearn Gesture dataset for body language understanding \cite{escalera2013multi} and the G3D dataset for gaming action analysis \cite{bloom2012g3d}.
More extensive empirical analysis of the proposed approach is also provided in this paper.
\section{The Proposed Method}
\label{sec:method}


We introduce the proposed network architecture, Scale Selection Network (SSNet), for skeleton-based online action prediction in this section.
The overall schema of this method is illustrated in \figurename{~\ref{fig:wavenet}}.
In the proposed network, the one dimensional (1-D) convolutions are performed in temporal domain to model the motion dynamics over the frames.
The inputs of SSNet are the frames within a temporal window at each time step.
In order to tackle the scale variations in the partially observed action at different time steps,
a scale selection method is proposed,
which enables our SSNet to focus on the observed part of the ongoing action by picking the most suitable convolutional layers.
To better deal with the input data modality,
a hierarchy of dilated tree convolutions are also introduced to process the input skeleton data for our network.

\subsection{Temporal Modeling with Convolutional Layers}
\label{sec:method:dilatedCNN}

\begin{figure}[tp]
	\centerline{\includegraphics[scale=0.39,trim={1.0cm 0.1cm 1.8cm 0.1cm},clip]{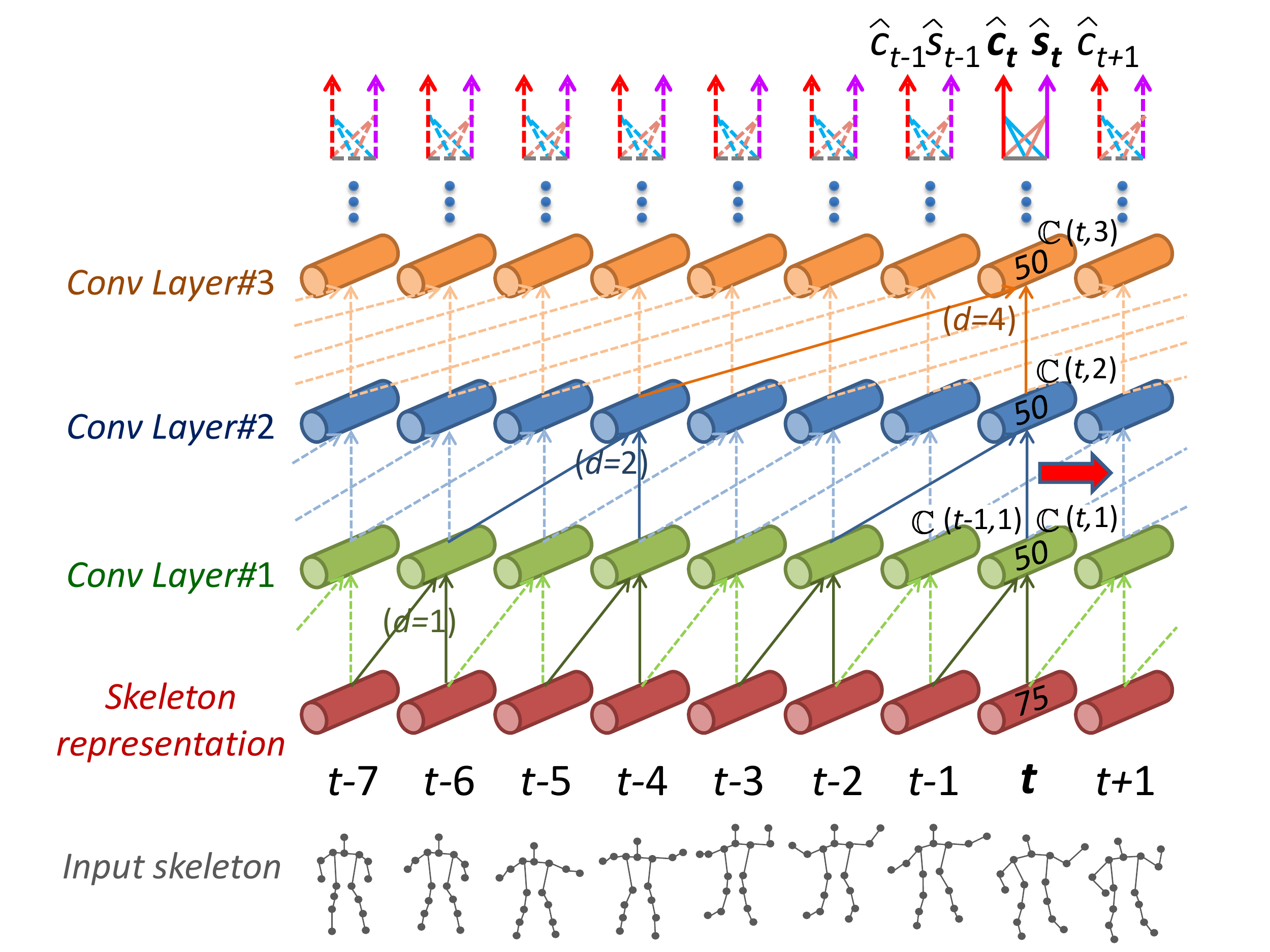}}
	\caption{
Illustration of the proposed SSNet for action prediction over the temporal axis.
The solid lines denote the SSNet links activated at current step $t$,
and the dashed lines indicate the links activated at other time steps.
Our SSNet has 14 1-D convolutional layers.
Here we only show 3 layers for clarity.
At each time step, SSNet predicts the class ($\hat{c}_t$) of the ongoing action,
and also estimates the temporal distance ($\hat{s}_t$) to current action's start point.
Calculation details of $\hat{c}_t$ and $\hat{s}_t$ are shown in \figurename{~\ref{fig:classreg}}.
Convolutional filters are shared at each layer, yet different across layers.
(Best viewed in color)
}
	\label{fig:wavenet}
\end{figure}

Convolutional networks \cite{lecun1995convolutional} have proven their superior strength in modeling the time series data \cite{van2016wavenet,dauphin2016language,varol2017long}.
For example, van den Oord \etal \cite{van2016wavenet} proposed a convolutional model, called WaveNet, 
for audio signal generation, and
Dauphin \etal \cite{dauphin2016language} introduced a convolutional network for time series in language sequential modeling. 
Inspired by the success of convolutional approaches in the analysis of temporal sequential data,
we leverage a stack of 1-D convolutional layers to model the motion dynamics and context dependencies over the video sequence frames,
and inspired by the WaveNet model, we propose a network for the skeleton-based action prediction task.
%
Specifically, a hierarchical architecture with dilated convolutional layers is leveraged in our model
to learn a comprehensive representation over the video frames within a temporal window. 

\begin{figure}[tp]
\centerline{\includegraphics[scale=0.39,trim={1.5cm 3cm 2.1cm 2.4cm},clip]{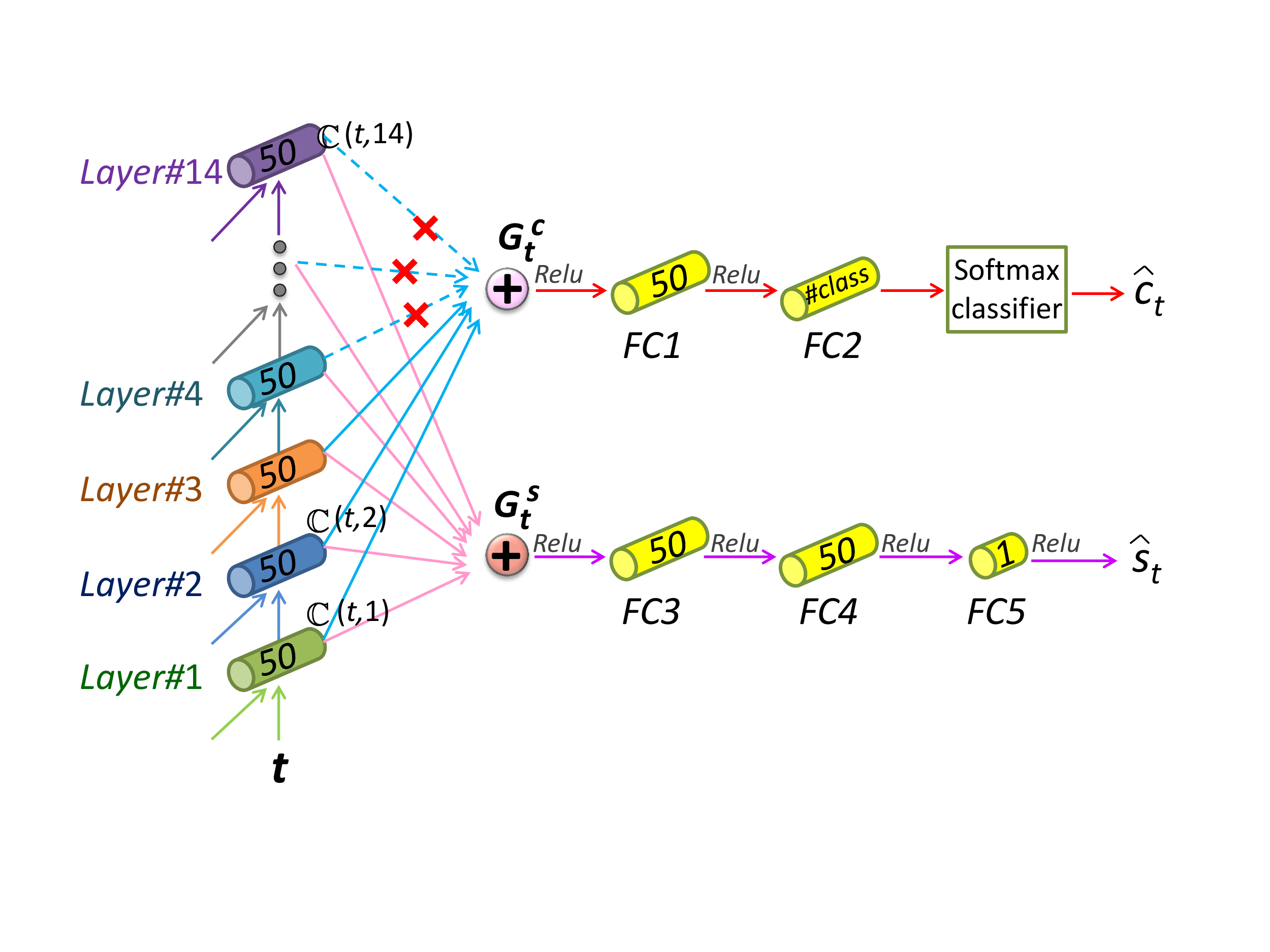}}
\caption{Details of our SSNet that jointly predicts the class label $\hat{c}_t$ and regresses the start point's distance $\hat{s}_t$ for the current ongoing action \textbf{at time \emph{t}}.
If the regressed result $\hat{s}_{t-1}$ at the previous time step ($t-1$) indicates that layer \#3 corresponds to the most \emph{proper} window scale (\ie, $l_t^p=3$),
then our network will use layers \#1-3 for class prediction,
while the activations from the layers above \#3 are dropped (marked with \emph{cross} in the figure).
In this figure, we only show a subset of convolutional nodes of our SSNet,
and other ones in the hierarchical structure (depicted as the solid lines in \figurename{~\ref{fig:wavenet}}) are omitted for clarity.
The parameters of the convolutional layers and FC (fully connected) layers in our SSNet are trained jointly in an end-to-end fashion.
}
\label{fig:classreg}
\end{figure}

\textbf{Dilated convolution.}
The main building blocks of our network model are dilated causal convolutions. 
Causal design \cite{van2016wavenet} enforces the prediction task at time $t$ to be based on the available information before $t$ (including $t$) without using the future information.
Dilated convolution \cite{YuKoltun2016} applies the convolutional filter over a larger field than the filter's length,
and some input values inside the field are skipped by a certain step size.

\begin{figure}[tp]
	\centerline{\includegraphics[scale=0.435,trim={0.2cm 5.8cm 4.93cm 3.9cm},clip]{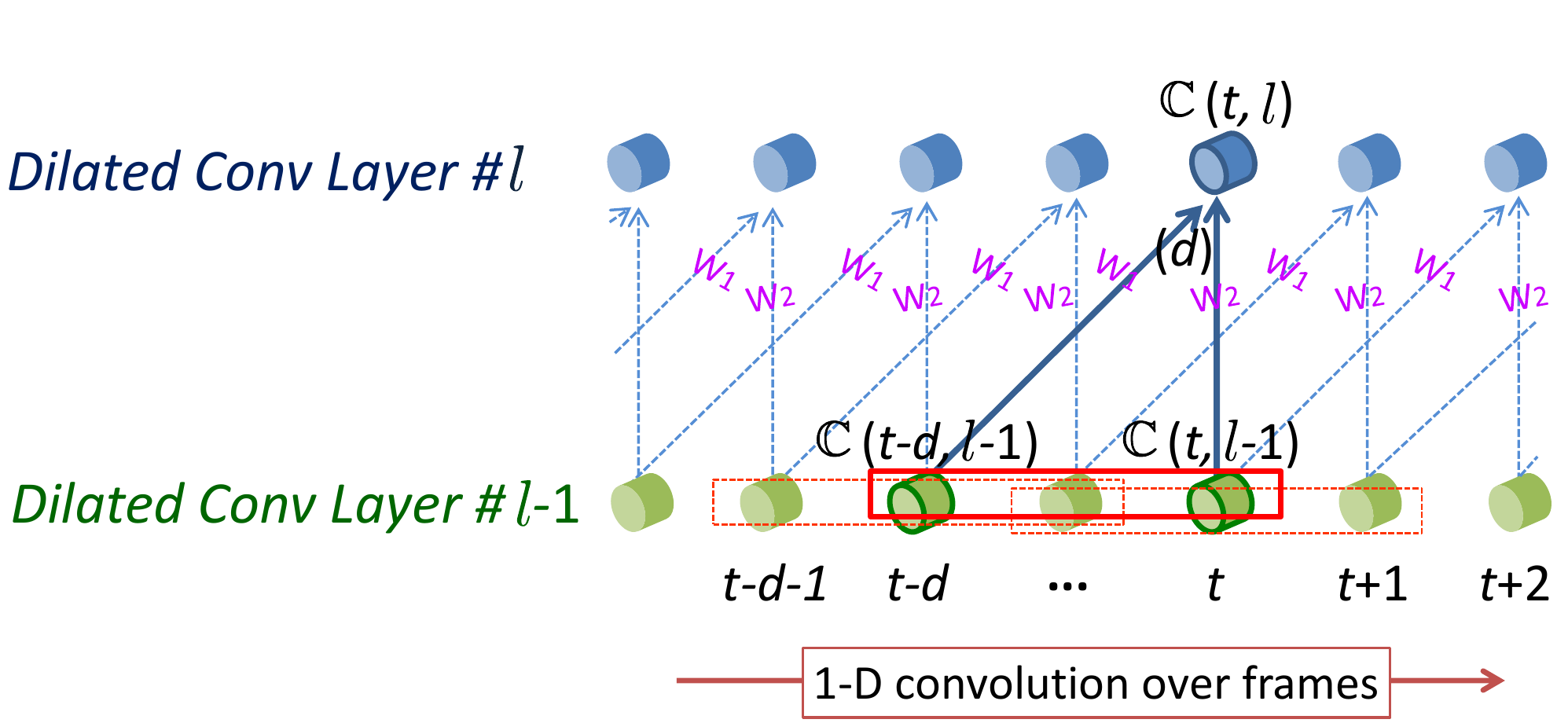}}
	\caption{Illustration of the dilated convolution layer used in our network.
            At each position, the 1-D convolutional filter covers a time range (labeled as a red box),
            and only the two boundary nodes (corresponding to two time steps) in the covered range are used, 
            while the other nodes between these two nodes are not used by the dilated convolutional operation for this position.
            }
	\label{fig:dilatedCausalConv}
\end{figure}

Concretely, dilated convolution (also known as ``convolution with holes'') can be formulated as presented in \cite{YuKoltun2016}:
\begin{eqnarray}
(X *_d w)(\mathbf{p}) ~ = \sum_{\mathbf{t} + d \mathbf{s} = \mathbf{p}} X(\mathbf{t}) ~ w(\mathbf{s})
\label{eq:dilatedConv}
\end{eqnarray}
where $*_d$ indicates the dilated convolutional operation, $X$ is the input, $w$ is the filter,
and $d$ denotes the dilation rate of the convolution ($d = 1$ represents the standard convolution).

In order to show how the dilated convolution is used in our model,
we illustrate the mechanism of a dilated convolutional layer in \figurename{~\ref{fig:dilatedCausalConv}}.

As shown in \figurename{~\ref{fig:dilatedCausalConv}},
at each position (\eg, the position $t$), the dilated convolutional filter (with dilation rate $d$) works over two input time steps ($t$ and $t-d$),
and the other time steps between these two steps are not considered for the convolutional operation at this position.
%
Let $\mathbb{C}(t,l)$ denote the activation of the convolutional node at the position $t$ in the dilated convolutional layer \#$l$
($l \in [1,\mathcal{L}]$, and $\mathcal{L}$ denotes the number of 1-D convolutional layers in our network).
Then $\mathbb{C}(t,l)$ can be calculated as:
\begin{eqnarray}
\mathbb{C}(t,l) = f \Big( W_{1} ~ \mathbb{C}(t-d,l-1) + W_{2} ~ \mathbb{C}(t,l-1) + b \Big)
\label{eq:C_tl}
\end{eqnarray}
where $f(\cdot)$ is a non-linear activation function.
$W_{1}$ and $W_{2}$ (together with the bias $b$) are the parameters of the dilated convolutional filter,
which are shared at the same layer, as illustrated in \figurename{~\ref{fig:dilatedCausalConv}}.

It is intuitive to use the aforementioned dilated convolution for human action analysis,
because the running time for longer actions can be very long and the convolutional network needs to be able to cover a large receptive field.
Applying standard convolution,
the network needs more layers or larger filter sizes to achieve a broader receptive field.
However, both of these significantly increase the number of model parameters.
In contrast,
by configuring the dilation rate ($d$),
dilated convolution can support expansion of the receptive field very efficiently, without bringing more parameters.
In addition, it does not need any extra pooling operations,
thus it can well maintain the ordering information of the inputs \cite{YuKoltun2016}.

\textbf{Multiple dilated convolutional layers.}
In our method, we stack multiple dilated convolutional layers, as illustrated in \figurename{~\ref{fig:wavenet}}.
%
The dilation rate increases exponentially over the layers in our network,
\ie, we set $d$ to $1,~2,~4,~8,~...$ for layers \#$1,~$\#$2,~$\#$3,~$\#$4,~...$, respectively.

This design results in an \textbf{exponential} expansion of the perception scale across the network layers.
For example,
the perception temporal window of the convolutional operation node $\mathbb{C}(t,2)$ in layer \#2 (see \figurename{~\ref{fig:wavenet}}) is $[t-3,t]$ ($4$ frames),
while the node $\mathbb{C}(t,3)$ in layer \#3 corresponds to a larger scale of temporal window ($8$ frames: $[t-7,t])$.


It is worth mentioning that 
all the video frames in the window $[t-7,t]$ can be perceived by the node $\mathbb{C}(t,3)$ with the hierarchical structure. 
This shows how the field of view expands over the layers in our network, while the coverage of the input is kept.


\subsection{Scale Selection}
\label{sec:method:SSS}

\begin{table*}[!tp]
\caption{Details of the main structure of SSNet. Refer to \figurename{~\ref{fig:overallarchit}} for the detailed architecture configurations of SSNet.}
\label{table:ssnetdetails}
\centering
\small
\begin{tabular}{ccccccccccccccc}
\toprule
1-D convolutional layer index                  & \#1 & \#2 & \#3  & \#4& \#5& \#6& \#7 & \#8 & \#9 &\#10 & \#11& \#12& \#13& \#14 \\
\midrule
Dilation rate ($d$)                          &   1 &   2 &   4  & 8  & 16 & 32 & 64  & 1   & 2   &   4 & 8   & 16  & 32  & 64 \\
Perception temporal window scale (frames)      &   2 &   4 &   8  & 16 & 32 & 64 & 128 & 129 & 131 & 135 & 143 & 159 & 191 & 255 \\
Output channels                                &  50 &  50 &  50  & 50 & 50 & 50 &  50 &  50 &  50 &  50 &  50 &  50 &  50 &  50 \\
\bottomrule
\end{tabular}
\end{table*}

For the streaming sequences,
we can utilize the frames in a temporal window $[t-s, t]$ (with scale $s$) to perform action prediction at the time step $t$.
However, finding a proper temporal scale $s$ for different steps and inputs is not easy.
At the early stages of an action,
a relatively small scale is preferred,
because larger windows can involve too many frames from the previous action, which may influence the recognition.
On the contrary, if a large ratio of the action is observed
(especially when the duration of this action is long),
to obtain a reliable prediction,
we need a larger $s$ to cover more of its observed parts.
This implies the importance of finding a proper scale value at each time step, rather than using a fixed scale at all steps.

We propose a scale selection scheme for online action prediction in this section. 
The core idea is to regress a \emph{proper} window scale at each time step,
and then at the next time step,
the network can use this scale value to choose the \emph{proper} layers for action prediction.

At each step, as shown in \figurename{~\ref{fig:wavenet}},
the class label ($\hat{c}_{t}$) of the current action is predicted,
and the temporal distance ($\hat{s}_t$) between the current action's start point and the current frame is also regressed.
This distance indicates that the performed part of the current action is assumed to be $[t-\hat{s}_t, t]$ at step $t$.

Assuming that we have obtained the regression result $\hat{s}_{t-1}$ at step $(t-1)$,
thus at frame $t$, our network selects the time range $[(t-1)-\hat{s}_{t-1}, t]$ for action prediction. 
Specifically, in our network design, the nodes in different layers correspond to different perception temporal window scales,
thus we can select the node from the \emph{proper} layer to cover the performed part of the current action.
For this \emph{proper} layer $l$, we make sure its perception window's scale equals to (or slightly larger than) $\hat{s}_{t-1}+1$,
while the perception window of its previous layer ($l-1$) is smaller than $\hat{s}_{t-1}+1$.
For example, layer \#2 in \figurename{~\ref{fig:coverfig}} is the \emph{proper} layer in this case.

Let $l_t^p$ denote the selected \emph{proper} layer at step $t$.
Then we aggregate the activations of the nodes $\mathbb{C}(t,l)$ ($l\in[1,l_t^p]$) in our network to generate a comprehensive representation for the selected time range as:
\begin{eqnarray}
\label{eq:skipconnectionC}
{G_t^c} = \frac{1}{l_t^p}\sum\limits_{l=1}^{l_t^p}  \mathbb{C}(t,l)
\end{eqnarray}
Note that we connect multiple layers ($[1,l_t^p]$) together to compute ${G_t^c}$,
rather than using $l_t^p$ only.
This skip connection design can speed up convergence and enables the training of much deeper models,
as shown by \cite{he2016deep,he2016identity}.
Besides, it can also help to improve the representation capability of our network,
as the information from multiple layers corresponding to multiple scales is fused for current action.
Finally, ${G_t^c}$ is fed to the fully connected layers followed by a softmax classifier to predict the class label ($\hat{c}_{t}$) for the current time step.

As shown in \figurename{~\ref{fig:classreg}}, beside predicting the action class ($\hat{c}_{t}$),
our network also generates a representation (${G_t^s}$) to regress the start point's distance ($\hat{s}_{t}$):
\begin{eqnarray}
\label{eq:skipconnectionS}
{G_t^s} = \frac{1}{\mathcal{L}}\sum\limits_{l=1}^{\mathcal{L}}  \mathbb{C}(t,l)
\end{eqnarray}


For the distance regression, we directly adopt the top convolutional layer $\mathcal{L}$ (\emph{together with all the layers below it}),
which has a large perception window (generally larger than the complete execution time of one action),
rather than dynamically selecting a layer as in Eq (\ref{eq:skipconnectionC}).
This is due to the essential difference between the regression task and the action label prediction task.
Start point's distance regression can be regarded as regressing the position of the bonding \cite{liu2017manifold} between the \emph{current action} and its previous activities,
thus involving information from the previous activity will not reduce (or even benefit) the regression performance for current action.
Using Eq (\ref{eq:skipconnectionS}) also implies
the distance regression is performed independently at each time step,
and is not affected by the regression results of the previous steps.

In the domain of object detection \cite{lin2017feature}, such as Fast-RCNN \cite{girshick2015fast}, the bounding box of the current object was shown to be accurately regressed by a learning scheme.
Similarly, our proposed network learns to regress the bounding (start point) of the current ongoing action reliably.

The regression result produced by the previous step ($t-1$) is used to guide the scale selection (with scale $\hat{s}_{t-1}+1$)
for action prediction at the current step $t$.
An alternative method can be: first regressing the scale $\hat{s}_{t}$ at step $t$,
then using the scale $\hat{s}_{t}$ to directly perform action prediction for the same step $t$.
We observe these two choices perform similarly in practice.
This is intuitive as $\hat{s}_{t-1}+1$ is close to $\hat{s}_{t}$.
The main difference of these two choices is the scale used at the beginning of a new action,
because if we use the scale regressed by its previous step, the scale used at this step may be derived from the previous action, which is not proper.
However, at the beginning frame of an action, too little information of the current action is observed,
which makes prediction at this step very difficult even using the proper scale (only one frame),
thus these two choices still perform similarly at this step.
In the following frames, since more information is observed and proper scales can be used, both choices perform reliably.
The framework will be less efficient if regressing for the same step,
as the two tasks (regression and prediction) need to be conducted as two sequential stages at each time step (cannot be performed simultaneously).

\subsection{Details of the Main Structure}

\begin{figure*}[tbp]
\centering
	\centerline{\includegraphics[scale=0.73,trim={0.1cm 0.01cm 0.01cm 0.3cm},clip]{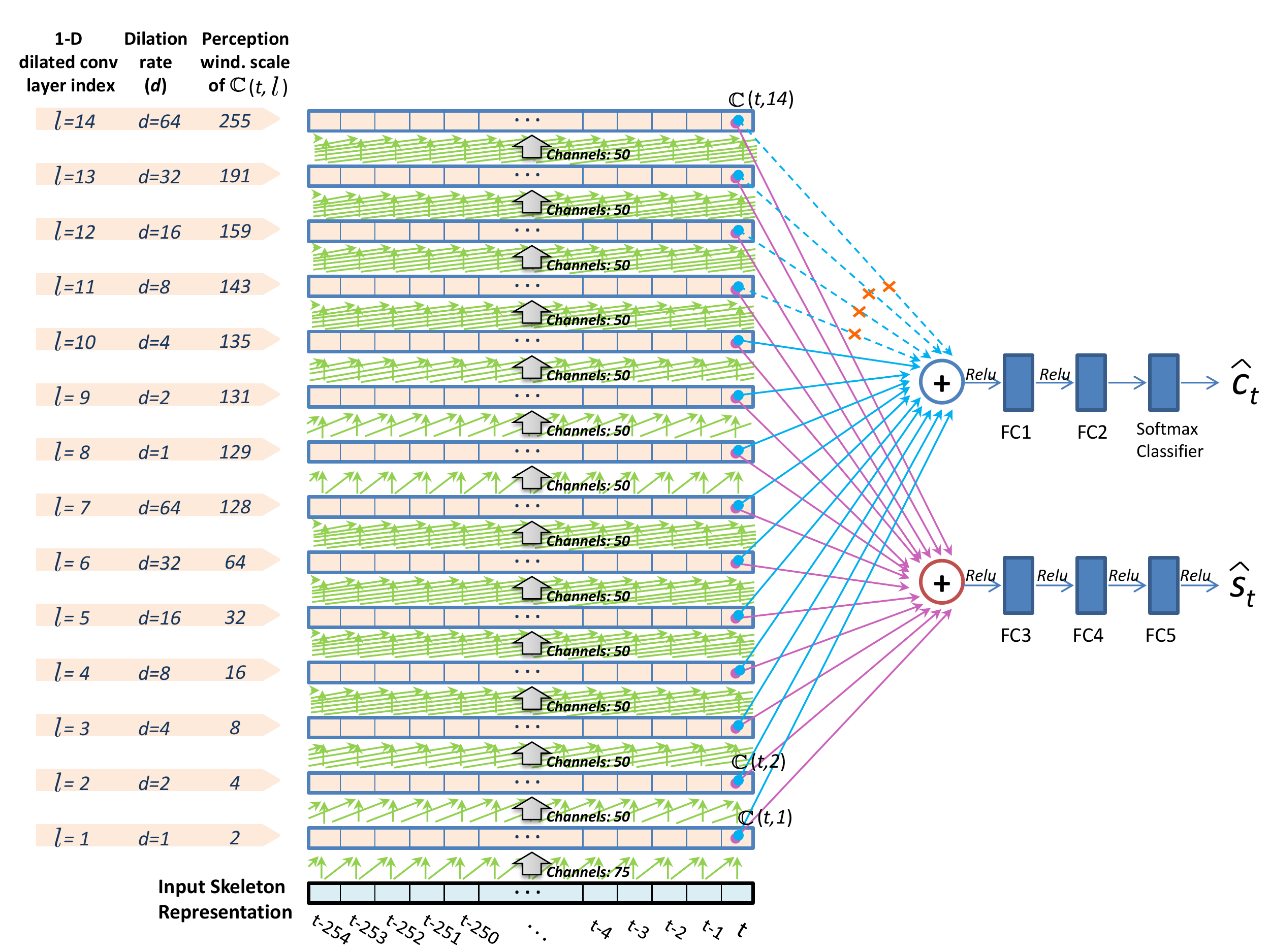}}
    \centering
    \caption{
        Detailed network architecture configurations of SSNet (for action prediction at the time step $t$).
        The distance regression is performed based on the top convolutional layer (together with the layers below it with the skip connections), which has a large perception window.
        The class prediction is performed based on the selected \emph{proper} layer (together with the layers below it), which is selected based on the estimated window scale.
    }
	\label{fig:overallarchit}
\end{figure*}

The proposed SSNet has 14 dilated convolutional layers for temporal modeling.
Specifically, we stack two similar sub-networks with dilation rates $(d):$ {$1,2,4,8,...,64$} over the layers of each sub-network,
\ie, the dilation rate $(d)$ is reset to 1 at the beginning of each sub-network, as shown in \tablename{~\ref{table:ssnetdetails}} and \figurename{~\ref{fig:overallarchit}}.
The motivation of this design is to achieve more variation for the temporal window scales (we obtain 14 different scales from 2 to 255 here).
Besides, each sub-network can be intuitively regarded and implemented as a large convolutional module.
Moreover, such a design still guarantees the node at each layer to perceive all the video frames in its perception window (\ie, without losing input coverage),
due to the hierarchial structure of SSNet.

With such a design, the perception temporal window scale of the top layer in our network is 255 frames,
which covers more than 8-second sequence at the recording frame rate of common video cameras like Kinect.
Generally, the duration of a full single action in most existing datasets is less than 8 seconds.
Thus, the temporal scale 255 is large enough for action analysis.
Even if the whole duration time of an action is longer than 8 seconds,
we believe the classification can be performed reliably when such a long segment (8 seconds) of the action has been perceived.



\subsection{Activation Sharing Scheme}
\label{sec:method:activationshare}


Our framework can be implemented in a very computation-efficient way.
Although both action label prediction and distance regression are conducted on various window scales at each step,
all of the computational steps are encapsulated in a single network with a hierarchical structure
(see \figurename{~\ref{fig:wavenet}}),
\ie, we do not need separated networks or multiple scanning passes for action prediction at each step. 

In addition, although convolutional operations are performed over a sliding window at each step,
the redundant computations of the overlapping regions among different sliding positions are avoided.
With the causal convolution design,
many features (activations of convolutional operations)
computed in previous steps can be reused by the latter steps,
which avoids redundant computation.

As depicted in Eqs (\ref{eq:skipconnectionC}) and (\ref{eq:skipconnectionS}),
at time step $t$, the prediction and regression are based on the nodes $\mathbb{C}(t,l)$,
$l\in[1,l_t^p]$ or $l\in[1,\mathcal{L}]$.
Each node $\mathbb{C}(t,l)$ is calculated based on only two input nodes,
$\mathbb{C}(t-d_{l},l-1)$ and $\mathbb{C}(t,l-1)$, as shown in \figurename{~\ref{fig:wavenet}}.
$\mathbb{C}(t-d_{l},l-1)$ has already been computed at time step $t-d_{l}$.
Therefore, to obtain $\mathbb{C}(t,l)$, we only need to calculate the activation of $\mathbb{C}(t,l-1)$.
Similarly, $\mathbb{C}(t,l-1)$ can be computed after we get $\mathbb{C}(t,l-2)$.

As a result, although we feed a window of frames to SSNet at each time step $(t)$,
we only need to calculate the activations of the nodes in column $t$ of \figurename{~\ref{fig:wavenet}},
and all other convolutional operations in the hierarchical structure can be copied from the previous time steps.
This activation sharing makes our network efficient enough to be used in real-time applications.

\subsection{Multi-level Structured Skeleton Representations}
\label{sec:method:treeCNN}

\begin{figure}[tp]
    \begin{minipage}[b]{0.45\linewidth}
		\centering
		\centerline{\includegraphics[scale=0.35,trim={6.9cm 2.9cm 8.0cm 2.0cm},clip]{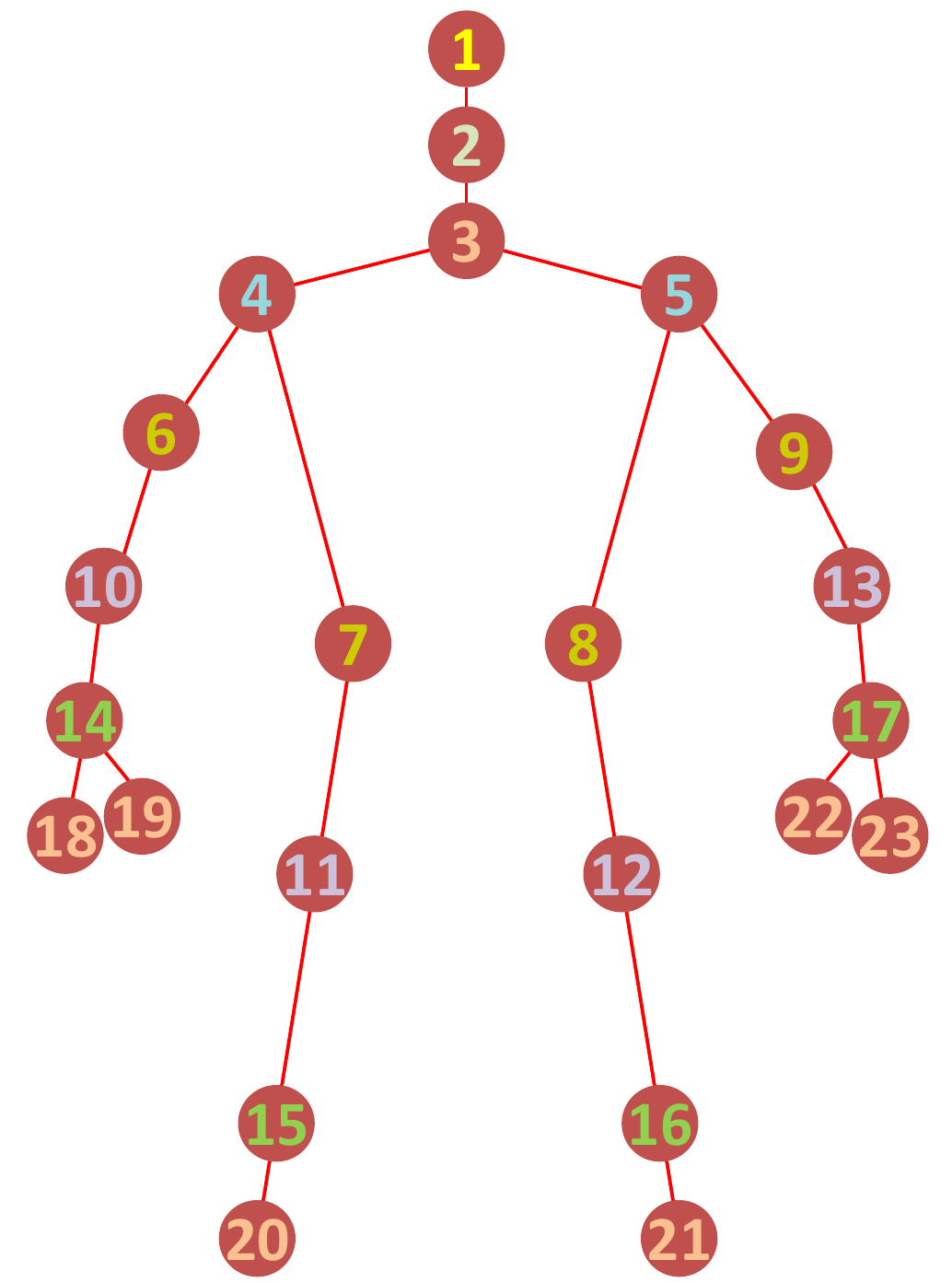}}
        (a)
	\end{minipage}
	\begin{minipage}[b]{0.549\linewidth}
		\centering
		\centerline{\includegraphics[scale=0.35,trim={6.3cm 4cm 6cm 5cm},clip]{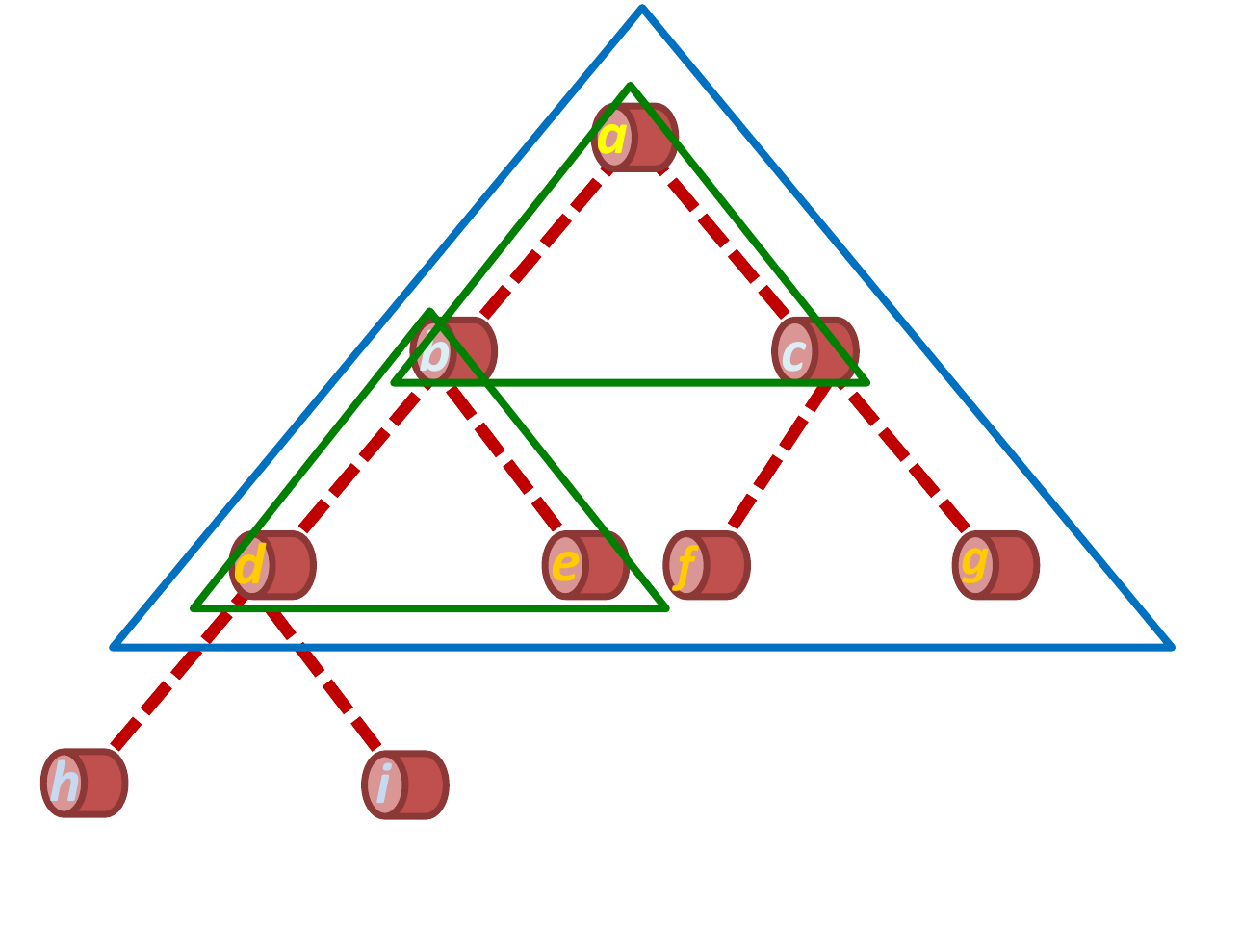}}
         (b)
	\end{minipage}
	\caption{
(a) The skeleton joints of the human body form a tree structure.
We set the head joint (joint $1$) as the root node,
and the height of the tree in this figure is $8$.
(b) Illustration of the convolution with triangular filters sliding over the tree structure.
The green and the blue triangles indicate the convolutions with two different filter sizes.
}
	\label{fig:skeleton_tree_strucutre}
\end{figure}

\begin{figure}[tp]
	\centerline{\includegraphics[scale=0.38,trim={1.8cm 0.1cm 2.8cm 3.3cm},clip]{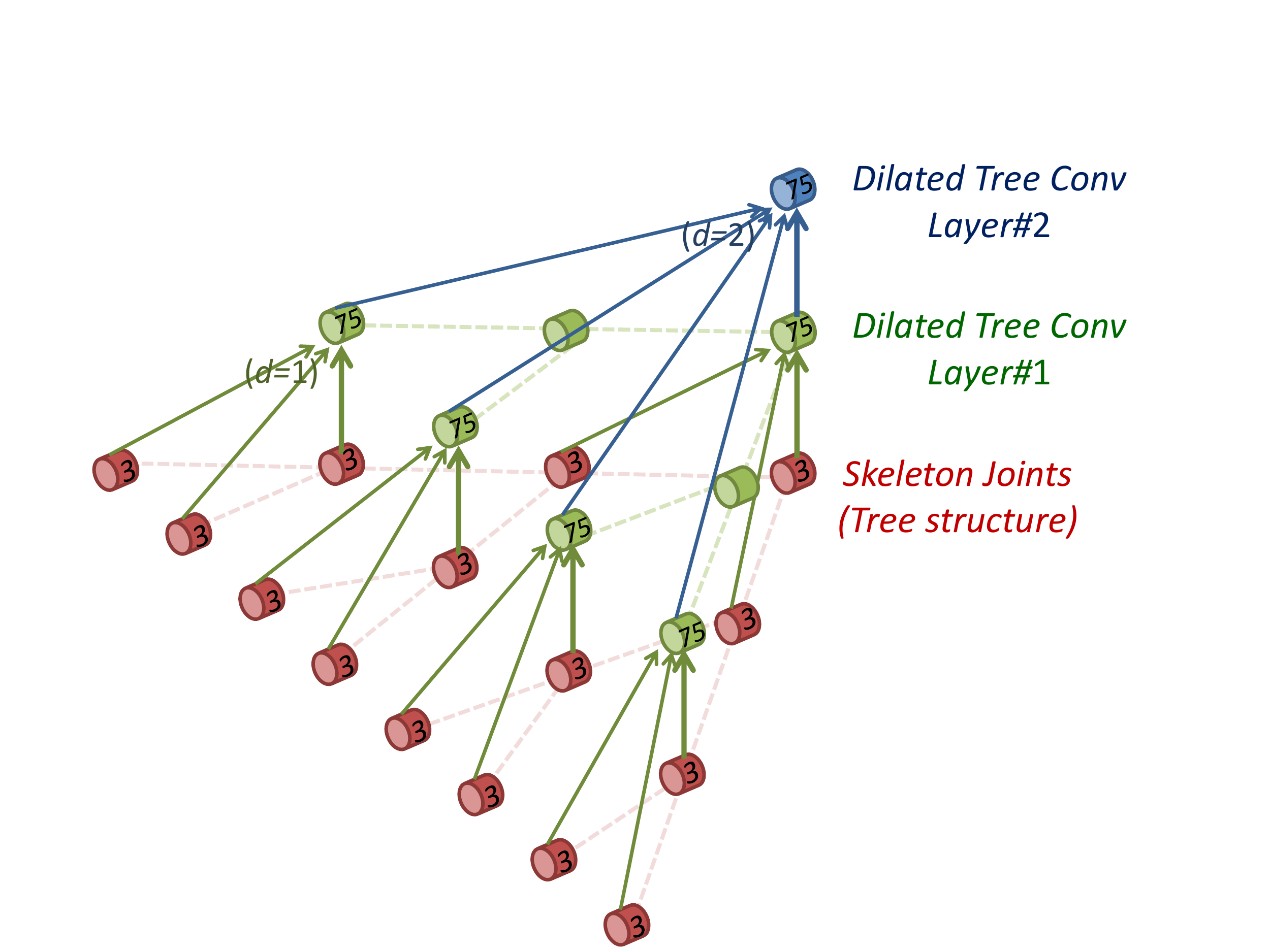}}
	\caption{
Illustration of the hierarchy of dilated tree convolutions that learns the multi-level structured representations over the input skeleton joints (labeled in red) at each frame.
The solid arrows denote the dilated tree convolutions with triangular filters. 
In our method, 3 dilated tree convolutional layers are used to cover the input skeleton tree with height 8,
while in this figure, we only show 2 layers that cover the tree with height 4 for clarity.
Note that the bottom of this figure shows a full binary tree,
while the human skeleton only has a subset of the nodes of a full binary tree.
Therefore, in implementation, the convolutional operations only need to be performed on a subset of the nodes.
The channel number of the input skeleton is 3, namely, the 3D coordinates $(x, y, z)$ of each joint.
(Best viewed in color)
}
	\label{fig:wavenetSkeleton}
\end{figure}

As mentioned above, in our framework,
the streaming 3D skeleton data is fed to the SSNet.
A naive way to perform action prediction with such an input data structure
is to concatenate the 3D coordinates of all joints at each frame to form a vector
(that we call it as coordinate concatenation representation).
We can then feed this coordinate concatenation representation of each frame to the SSNet as input (see \figurename{~\ref{fig:wavenet}}).
However, the semantic structure amongst the skeleton joints in a frame is ignored in this representation.
As illustrated in \figurename{~\ref{fig:skeleton_tree_strucutre}(a)},
the skeleton joints in every human body configuration are physically connected in a semantical tree structure in the spatial domain,
and utilizing such structure information has shown to be quite helpful for human activity analysis \cite{du2015hierarchical,liu2016eccv, shahroudy2016multimodal}.

Instead of directly using the method of coordinate concatenation,
in this paper,
we model the spatial tree structure of the skeleton joints,
in order to capture the posture information of the human body more effectively at each frame
and thus strengthen the capability of our framework in skeleton-based action prediction.

Specifically, we propose a hierarchy of dilated tree convolutions in spatial domain 
to learn the multi-level (local, mid-level, and holistic) structured representations for the tree structure of the skeleton in each frame.
The proposed hierarchical dilated tree convolution for spatial domain modeling is essentially
an extension of the multi-layer 1-D dilated convolution that is introduced in section \ref{sec:method:dilatedCNN} for temporal modeling.
Below we introduce this design in detail.





\textbf{Convolution over tree structure.}
Convolutional networks are powerful tools in modeling the spatial visual structures \cite{lecun1995convolutional}.
Here to model the discussed semantic structure of the human skeleton,
we propose to apply convolutions 
by using triangular filters sliding over the nodes of the tree, as shown in \figurename{~\ref{fig:skeleton_tree_strucutre}(b)}.
At each step of the convolution, the triangular filter covers a sub-tree region,
and the nodes in this region are used to produce an activation as a semantic representation of this position.
This process is similar to the common convolutional operations that slide over the pixels of an input image or previous layer's feature maps.
Different sizes of the triangular filters can also be used for this process,
as shown in \figurename{~\ref{fig:skeleton_tree_strucutre}}(b).

In our method,
zero padding is adopted for the convolution over the skeleton tree,
\ie, if a certain node (\eg, joint 2 in \figurename{~\ref{fig:skeleton_tree_strucutre}(a)}) has only one child (joint 3),
to perform convolution at this node position,
we set this child (joint 3) as the left node, and its right node is filled with zero.
Similarly, for the leaf nodes (\eg, joint 20), both of the child nodes are filled with zero.

\textbf{A hierarchy of dilated tree convolutions.}
In order to learn representations that are effective and discriminative for representing the skeletal data in a frame,
we stack multiple convolutional layers over the tree-structured skeleton joints,
and perform convolution with triangular filters at each layer,
as illustrated in \figurename{~\ref{fig:wavenetSkeleton}}.

Dilated convolutions which are effective and efficient in computation are also used here (similar to section \ref{sec:method:dilatedCNN}).
Only the top and the bottom nodes in each triangular region of each position are used for activation calculation,
as shown by the Layer \#1 (with dilation rate set to 1) and Layer \#2 (with dilation rate set to 2) in \figurename{~\ref{fig:wavenetSkeleton}}.
Here we call this convolution design as dilated tree convolution.

Three dilated tree convolutional layers are stacked in our model,
and their dilation rates are 1, 2, and 4, respectively.
Therefore, a hierarchy of dilated tree convolutions are constructed over the skeletal data.
The details of this hierarchy design are shown in \tablename{~\ref{table:wavenetSkeletonDetails}}.

With this design,
the nodes in different layers of the hierarchy perceive different spatial ranges of the input skeleton joints.
For example, \emph{each node} in Layer \#1 of \figurename{~\ref{fig:wavenetSkeleton}} learns a representation from a very local region of neighbouring joints of the input skeleton
(perception sub-tree height is 2),
while \emph{each node} in Layer \#2 learns a representation over a larger region of the skeleton
(perception sub-tree height is 4).
Specifically,
the top layer, \#3, can learn a representation based on all the joints of the whole skeleton tree
(perception tree height is 8).
This implies that the multi-level (local, mid-level, and holistic) structured semantic representations of the skeleton data are learned at different layers in this hierarchy.

Finally, we aggregate the multi-level representations by averaging the activations of all the convolutional nodes in the hierarchy,
and the aggregated result is fed to our SSNet as the representation of the skeleton data at each frame (see \figurename{~\ref{fig:wavenet}}).

Since the multi-level structured semantic representations are learned,
which are effective for representing the spatial structure and posture of the human skeleton at each frame,
the performance of our SSNet for action prediction is improved.
Moreover, this structured skeleton representation learning procedure can be attached to our SSNet as an input processing module of it (see \figurename{~\ref{fig:wavenet}}),
such that the whole model of our SSNet is still end-to-end trainable.

\begin{table}[!tp]
\caption{Details of the hierarchy of dilated tree convolutions (corresponding to \figurename{~\ref{fig:wavenetSkeleton}}).}
\label{table:wavenetSkeletonDetails}
\centering
\small
\begin{tabular}{cccc}
\toprule
Dilated tree convolutional layer index   &   ~\#1~  &   ~\#2~ & ~\#3~   \\
\midrule
Dilation rate ($d$)                    &     1    &     2   &   4  \\
Perception sub-tree height                   &     2    &     4   &   8   \\
Output channels                          &    75    &     75  &  75   \\
\bottomrule
\end{tabular}
\end{table}

\subsection{Objective Function}

The objective function of our SSNet is formulated as:
\begin{eqnarray}
\label{eq:objfunc}
\ell= \ell_{c}(\hat{c}_{t},c_t) + \gamma \ell_{s}(\hat{s}_{t},s_t)
\end{eqnarray}
where $c_t$ is the ground truth class label, and $s_t$ is the ground truth distance between the start point of the action and the current frame $t$.
$\gamma$ is the weight for the regression task.
$\ell_{c}$ is the negative log-likelihood loss measuring the difference between the true class label $c_t$ and the predicted result $\hat{c}_{t}$ at time step $t$.
$\ell_{s}$ is the regression loss defined as $\ell_{s}(\hat{s}_{t},s_t)=(\hat{s}_{t}-s_t)^2$.
Our objective function is minimized by stochastic gradient descent.

To train our SSNet, we generate fixed-length clips from the annotated long sequences with sliding temporal windows.
The length of each clip is equal to the perception temporal scale of the top convolutional layer (255 frames).
Each clip can then be fed to the SSNet.
In the training phase,
class prediction is performed using the proper layer that is chosen based on the ground truth distance to the start point.
We also observe adding small random noise to the layer choosing process during training is helpful for improving the generalization capability of our network for class prediction.

In the testing phase, the action prediction is performed frame-by-frame through a sliding window,
and the proper layer for prediction at each time step is determined by the distance regression result of its previous step.
The ground truth information of the start point is not used during testing.


%

\section{Experiments}
\label{sec:experiments}


The proposed method is evaluated on four challenging datasets:
the OAD dataset \cite{li2016online}, the ChaLearn Gesture dataset \cite{escalera2013multi}, the PKUMMD dataset \cite{liu2017pku}, and the G3D dataset \cite{bloom2012g3d}.
In all the datasets, multiple action instances are contained in each long video.
Beside the predefined action classes,
these datasets also contain frames which belong to the background activity,
thus we add a blank class to represent the frames in this situation. 
We conduct extensive experiments with the following different architectures:

\begin{enumerate}
  \item
\textbf{SSNet.}
This is our proposed network for skeleton-based action prediction,
which can select a proper layer to cover the performed part of the current ongoing action at each time step \emph{by using the start point regression result}.
The multi-level structured skeleton representations are used in this network.


  \item
\textbf{FSNet ($\mathcal{S}$).}
Fixed Scale Network (FSNet) is similar to SSNet,
but the action prediction is directly performed using the top layer.
This indicates scale selection scheme is not used,
and the prediction is based on a fixed window scale ($\mathcal{S}$) at all steps.
We configure the structure and propose a set of FSNets,
such that they have different perception window scales at the top layer.
Concretely, five FSNets with different fixed scales ($\mathcal{S} = 15,~31,~63,~127,~255$) are evaluated.
To make a fair comparison,
skip connections (see Eq (\ref{eq:skipconnectionC})) are also used in each FSNet,
\ie, all layers (corresponding to different scales) in a FSNet are connected as Eq (\ref{eq:skipconnectionC}) for action prediction at each step.
  \item
\textbf{FSNet-MultiNet.}
This baseline is a combination of multiple FSNets.
A set of FSNets with different scales ($\mathcal{S} = 15,~31,~63,~127,~255$) are used for each time step.
We then fuse the results of them, 
\ie, exhaustive multi-scale multi-round scans are used to perform action prediction at each time step.
  \item
\textbf{\emph{SSNet-GT}.}
Beside the aforementioned models, we also evaluate an ``ideal'' baseline, \emph{SSNet-GT}.
Action prediction in \emph{SSNet-GT} is also performed at the selected layer.
However, we do not use the regression result to select the scale,
instead, we directly use the \emph{ground truth (GT)} distance of the start point to select the layer for action prediction at each step.
\end{enumerate}

Note that the multi-level structured skeleton representations are used
in all of the above architectures (SSNet, FSNet ($\mathcal{S}$), FSNet-MultiNet, and SSNet-GT) for fair comparisons.

Our proposed approach is also compared to other state-of-the-art methods for skeleton-based activity analysis:

\begin{enumerate}
  \item
\textbf{ST-LSTM} \cite{liu2017PAMI}.
This network achieves superior performance on 3D skeleton-based action recognition task.
We adapt it to our online action prediction task and generate a prediction of the action class at each frame of the streaming sequence.
  \item
\textbf{JCR-RNN} \cite{li2016online}.
This network is a variant of LSTM, which models the context dependencies in temporal dimension of the untrimmed sequences. 
It obtains state-of-the-art performance of action detection in skeleton sequences on some benchmark datasets.
A prediction of the current action class is provided at each frame of the streaming sequence.
  \item
\textbf{Attention Net} \cite{liu2017global}.
This network adopts an attention mechanism to dynamically assign weights to different frames and different skeletal joints for 3D skeleton-based action classification.
A prediction of the action class is produced at each time step.
\end{enumerate}


\subsection{Implementation Details}

%


The experiments are conducted with the Torch7 toolbox \cite{collobert2011torch7}.
Our network is trained from scratch,
\ie, the network parameters are initialized with small random values (uniform distribution in [-0.08, 0.08]).
The learning rate, momentum, and decay rate are set to $10^{-3}$, $0.9$, and $0.95$, respectively.
The output dimensions of FC1, FC3, FC4, and FC5 in \figurename{~\ref{fig:classreg}} are set to $50$, $50$, $50$, and $1$, respectively.
FC2's output dimension is determined by the class number of each specific dataset.
GLU \cite{dauphin2016language} is the activation function used for the convolutional operations in our network (see Eq (\ref{eq:C_tl})).
Residual connections \cite{he2016deep} are used over different convolutional layers.
The output channels of the convolutional nodes for temporal modeling (see \figurename{~\ref{fig:wavenet}}) are all 50.
The output channels of the convolutional nodes for structured skeleton representation learning are equal to the dimension of the coordinate concatenation representation of a frame.
In our experiment, $\gamma$ in Eq (\ref{eq:objfunc}) is set to $0.01$.
The above-mentioned parameters are obtained by cross-validation on the training sets.

In our SSNet, the proposed hierarchy of dilated tree convolution is used to learn the multi-level structured representation for each skeleton in a frame.
If two skeletons are contained in a frame, then their structured representations are averaged.
The averaged result is used as the representation for this frame.


We show the number of parameters of our SSNet with the two different skeleton representations in \tablename{~\ref{table:paraAndSpeed}}.
By attaching the multi-level structured representation,
the parameter number in the whole model of SSNet is only slightly larger than the configuration in which we use coordinate concatenation representation.
This implies that the number of parameters in the hierarchy of dilated tree convolutions is quite small (only $13\%$ of the whole model).

We also summarize the numbers of network parameters for different methods.
The numbers of network parameters of
SSNet,  FSNet(15), FSNet(31), FSNet(63), FSNet(127), FSNet(255), FSNet-MultiNet,  SSNet-GT, ST-LSTM, JCR-RNN, and AttentionNet are
310K,   170K,      200K,      240K,      270K,       310K,       1M,              310K,     420K,    290K,    and 3M,      respectively.

We perform our experiments with a single NVIDIA TitanX GPU.
We evaluate the efficiency of our method for online action prediction in the streaming sequence, and show the running speed of it in \tablename{~\ref{table:paraAndSpeed}}.
Our network responds fast for online action prediction. 
The low computational cost of our method is partially due to (1) the concise skeleton data as input, (2) the efficient dilated convolution, and (3) our activation sharing scheme.
Besides, even if we learn the multi-level structured representations, the overall speed of our SSNet is still very fast.

\begin{table}[h]
\caption{
Number of parameters and computational efficiency of our SSNet when using different skeleton representations within it.
}
\label{table:paraAndSpeed}
\centering
\begin{tabular}{ccc}
\toprule
Skeleton representations in SSNet         & \#Parameters      & Speed    \\
\midrule
With coordinate concatenation                 &  $270K$           &  $50fps$     \\
With multi-level structured representation    &  $310K$           &  $40fps$       \\
\bottomrule
\end{tabular}
\end{table}

\subsection{Experiments on the OAD Dataset}

The OAD dataset \cite{li2016online} was collected with Kinect v2 in daily-life indoor environments.
Ten action classes were performed by different subjects.
The long video sequences in this dataset correspond to about 700 action instances. 
The starting and ending frames of each action are annotated in this dataset.
In this dataset,
30 long sequences are used for training, and 20 long sequences are used for testing.

\begin{figure}[htbp]
		\centering
		\centerline{\includegraphics[trim=290 268 290 268,scale=0.5]{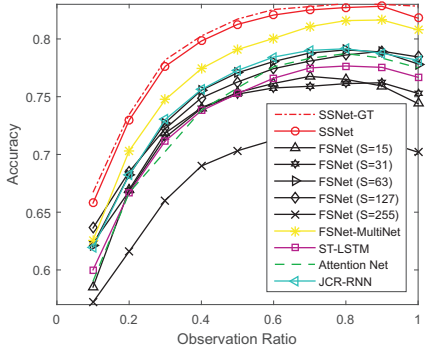}}
	\caption{Action prediction results on the OAD dataset.}
	\label{fig:result_curves_OAD}
\end{figure}

The action prediction results on the OAD dataset are shown in \figurename{~\ref{fig:result_curves_OAD}} and \tablename{~\ref{table:resultOAD}}.
In the figures and tables,
the prediction accuracy of an observation ratio $p\%$ denotes the average accuracy of the
predictions in the observed segment ($p\%$) of the action instance.  

\begin{table}[htbp]
	\caption{Action prediction accuracies on the OAD dataset.
        Note that in the last row, \emph{SSNet-GT} is an ``ideal'' baseline,
		in which the \emph{ground truth (GT)} scales are used for action prediction.
		Our SSNet, which performs prediction with the regressed scales, is even comparable to \emph{SSNet-GT}.
        Refer to \figurename{~\ref{fig:result_curves_OAD}} for more results.
	}
	\label{table:resultOAD}
	\centering
    \small
	\begin{tabular}{lccc}
    \toprule
    Observation Ratio                    & 10\% & 50\% & 90\% \\
    \midrule 
    JCR-RNN                              & 62.0\% & 77.3\% & 78.8\% \\
    ST-LSTM                              & 60.0\% & 75.3\% & 77.5\% \\
    Attention Net                        & 59.0\% & 75.8\% & 78.3\% \\
    \midrule
    FSNet (15)                           & 58.5\% & 75.4\% & 75.9\% \\
    FSNet (31)                           & 62.3\% & 75.2\% & 76.2\% \\
    FSNet (63)                           & 62.2\% & 77.1\% & 78.9\% \\
    FSNet (127)                          & 63.6\% & 76.3\% & 78.9\% \\
    FSNet (255)                          & 57.2\% & 70.3\% & 71.2\% \\
    FSNet-MultiNet                       & 62.6\% & 79.1\% & 81.6\% \\
    \midrule
    SSNet                                & \textbf{65.8}\%& \textbf{81.3}\% & \textbf{82.8}\% \\
    \midrule
    \emph{SSNet-GT}                      & \emph{66.7}\% & \emph{81.7}\% & \emph{83.0}\% \\
    \bottomrule
	\end{tabular}
\end{table}

Note that the special baseline \emph{SSNet-GT} performs action prediction with the \emph{ground truth} scale at each step, thus it provides the best results.
Our SSNet with regressed scale even achieves comparable results to this ``ideal'' baseline (\emph{SSNet-GT}),
which indicates the effectiveness of our scale selection scheme for online action prediction at each progress level.

Apart from the ``ideal'' \emph{SSNet-GT} model,
our proposed SSNet yields the best prediction results among all methods at all observation ratios.
Specifically, our SSNet can even produce a quite reliable prediction (about $66\%$ accuracy) at the early stage when only a small ratio ($10\%$) of the action instance is observed.


The performance of our SSNet is much better than FSNets which perform prediction with fixed-scale windows at each time step.
Even fusing a set of FSNets with different scales,
FSNet-MultiNet is still weaker than our single SSNet at all progress levels.
This demonstrates that our proposed scale selection scheme,
which guides the SSNet to dynamically cover the performed part of the current action at each step,
is very effective for online action prediction.

The proposed SSNet significantly outperforms the state-of-the-art RNN/LSTM based methods, JCR-RNN \cite{li2016online} and ST-LSTM \cite{liu2017PAMI},
which can handle continuous streaming skeleton sequences. 
The performance disparity could be explained as:
(1) At the early stages (eg. $10\%$), our SSNet can focus on the performed part of current action by using the selected scale,
while RNN models \cite{li2016online,liu2017PAMI} may bring information from the previous actions which can interfere the prediction for current action.
(2) At the latter stages (eg. $90\%$), the context information from the early part of current action may vanish in RNN model with its hidden state evolving frame by frame,
while our SSNet, which uses convolutional layers to model the temporal dependencies over the frames, can still handle the long-term context dependency information in the temporal window.
Our SSNet also outperforms the Attention Net \cite{liu2017global} that assigns weights to differen frames and joints.
This indicates the superiority of our SSNet with explicit scale selection.

We also observe the average action prediction accuracy decreases at the ending stages.
A possible explantation is that the frames at the ending stages of some action instances contain postures and motions that are not very relevant to the current action's class label.

\subsection{Experiments on the ChaLearn Gesture Dataset}

The ChaLearn Gesture dataset \cite{escalera2013multi} is a large-scale dataset for human action (body language) analysis,
which consists of 23 hours of Kinect videos.
A total of 20 action classes were performed by 27 subjects.
This dataset is very challenging, as the body motions of many action classes are very similar.

Unlike the NTU RGB+D dataset \cite{Shahroudy_2016_CVPR}, in which every video contains only one action,
each video in the ChaLearn Gesture dataset includes multiple (8$\sim$20) action instances.
Thus this dataset is suitable for online action prediction.
The starting and ending frames of $11116$ action instances are annotated.
On this dataset, $3/4$ of the annotated videos are used for training,
and the remaining annotated videos are held for testing.
We sample 1 frame from every 4 frames considering the large amount of data.

\begin{figure}[htbp]
		\centering
		\centerline{\includegraphics[trim=290 268 290 268,scale=0.5]{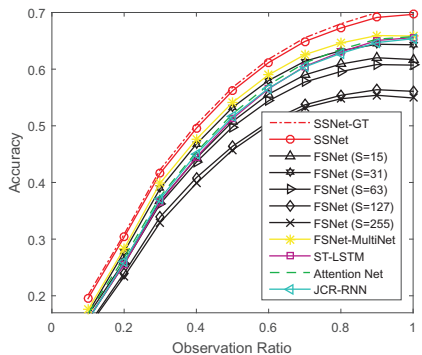}}
	\caption{Action prediction results on the ChaLearn Gesutre dataset.}
	\label{fig:result_curves_ChaLearn}
\end{figure}

We report the action prediction results in \figurename{~\ref{fig:result_curves_ChaLearn}} and \tablename{~\ref{table:resultChaLearn}}.
Our SSNet outperforms other methods at all observation ratios on this large-scale dataset.

\begin{table}[htbp]
	\caption{Action prediction accuracies on the ChaLearn Gesture dataset.
        Refer to \figurename{~\ref{fig:result_curves_ChaLearn}} for more results.
	}
	\label{table:resultChaLearn}
	\centering
    \small
	\begin{tabular}{lccc}
    \toprule
    Observation Ratio                    &  10\%  & 50\%   & 90\% \\
    \midrule
    JCR-RNN                              & 15.6\% & 51.6\% & 64.7\% \\
    ST-LSTM                              & 15.8\% & 51.3\% & 65.1\% \\
    Attention Net                        & 16.8\% & 52.1\% & 65.3\% \\
    \midrule
    FSNet (15)                           & 16.6\% & 50.8\% & 62.0\% \\
    FSNet (31)                           & 16.9\% & 53.2\% & 64.4\% \\
    FSNet (63)                           & 15.8\% & 49.8\% & 60.8\% \\
    FSNet (127)                          & 14.8\% & 46.4\% & 56.4\% \\
    FSNet (255)                          & 14.5\% & 45.7\% & 55.4\% \\
    FSNet-MultiNet                       & 17.5\% & 54.1\% & 65.9\% \\
    \midrule
    SSNet                                & \textbf{19.5}\%& \textbf{56.2}\% & \textbf{69.1}\% \\
    \midrule
    \emph{SSNet-GT}                      & \emph{20.1}\% & \emph{56.8}\% & \emph{70.0}\% \\
    \bottomrule
	\end{tabular}
\end{table}

\subsection{Experiments on the PKUMMD Dataset}

\begin{figure}[htbp]
		\centering
		\centerline{\includegraphics[trim=290 268 290 268,scale=0.5]{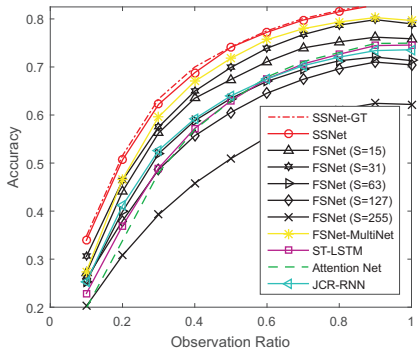}}
	\caption{Action prediction results on the PKUMMD dataset.}
	\label{fig:result_curves_PKUMMD}
\end{figure}

The PKUMMD dataset \cite{liu2017pku} was captured for RGBD-based activity analysis in continuous sequences.
Cross-subject evaluation protocol is used for this dataset,
in which 57 subjects are used for training, and the remaining 9 subjects are for testing.
Considering the large amount of data, we use the videos which contain the challenging interaction actions for our experiment, and sample 1 frame from every 4 frames for these videos.
The comparison results of the prediction performance on this dataset are presented in \figurename{~\ref{fig:result_curves_PKUMMD}} and \tablename{~\ref{table:resultPKUMMD}}.

Our method achieves the best results at all the progress levels on this dataset.
Specifically, our SSNet outperforms other methods significantly, even when only a very small ratio (10\%) of the action is observed.
This indicates that our method can produce a much better prediction at the early stage by focusing on the current action, compared to other methods which do not explicitly consider the scale selection.

Another observation is that the FSNet with fixed scale at each time step is quite sensitive to the scale used,
as different scales provide very different results.
This further demonstrates that our SSNet, which dynamically chooses the proper scale at each step to perform prediction, is effective for online action prediction.

\begin{table}[htbp]
	\caption{Action prediction accuracies on the PKUMMD dataset.
     Refer to \figurename{~\ref{fig:result_curves_PKUMMD}} for more results.
	}
	\label{table:resultPKUMMD}
	\centering
    \small
	\begin{tabular}{lccc}
    \toprule
    Observation Ratio                    & 10\%   & 50\%   & 90\% \\
    \midrule
    JCR-RNN                              & 25.3\% & 64.0\% & 73.4\% \\
    ST-LSTM                              & 22.9\% & 63.0\% & 74.5\% \\
    Attention Net                        & 19.8\% & 62.9\% & 74.9\% \\
    \midrule
    FSNet (15)                           & 27.1\% & 67.4\% & 76.2\% \\
    FSNet (31)                           & 30.6\% & 69.9\% & 79.8\% \\
    FSNet (63)                           & 25.3\% & 63.5\% & 72.1\% \\
    FSNet (127)                          & 25.9\% & 60.6\% & 71.0\% \\
    FSNet (255)                          & 20.2\% & 50.9\% & 62.4\% \\
    FSNet-MultiNet                       & 27.4\% & 71.8\% & 80.3\% \\
    \midrule
    SSNet                                & \textbf{33.9}\%& \textbf{74.1}\% & \textbf{82.9}\% \\
    \midrule
    \emph{SSNet-GT}                      & \emph{34.8}\% & \emph{74.2}\% & \emph{83.1}\% \\
    \bottomrule
	\end{tabular}
\end{table}

\subsection{Experiments on the G3D Dataset}

\begin{figure}[htbp]
		\centering
		\centerline{\includegraphics[trim=290 268 290 268,scale=0.5]{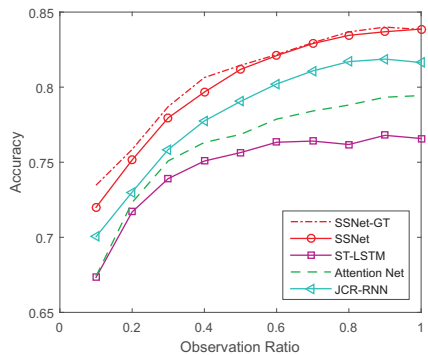}}
	\caption{Action prediction results on the G3D dataset.}
	\label{fig:result_curves_G3D}
\end{figure}

The G3D dataset \cite{bloom2012g3d} containing 20 gaming actions was collected with a Kinect camera.
There are 209 untrimmed long videos in this dataset.
We use 104 videos for training, and the remaining ones are used for testing.
Our SSNet achieves superior performance on this challenging dataset, as shown in \figurename{~\ref{fig:result_curves_G3D}} and \tablename{~\ref{table:resultG3D}}.

\begin{table}[htbp]
	\caption{Action prediction accuracies on the G3D dataset.
        Refer to \figurename{~\ref{fig:result_curves_G3D}} for more results.
	}
	\label{table:resultG3D}
	\centering
    \small
	\begin{tabular}{lccc}
    \toprule
    Observation Ratio                    & 10\%   & 50\%   & 90\% \\
    \midrule
    JCR-RNN                              & 70.0\% & 79.1\% & 81.9\% \\
    ST-LSTM                              & 67.3\% & 75.6\% & 76.8\% \\
    Attention Net                        & 67.4\% & 76.9\% & 79.3\% \\
    \midrule
    SSNet                                & \textbf{72.0}\%& \textbf{81.2}\% & \textbf{83.7}\% \\
    \midrule
    \emph{SSNet-GT}                      & \emph{73.5}\% & \emph{81.5}\% & \emph{84.0}\% \\
    \bottomrule
	\end{tabular}
\end{table}

\subsection{Evaluation of Skeleton Representations}
\label{sec:experiment:skeRep}

We compare the performance of our SSNet when using the multi-level structured skeleton representations to that when using the coordinate concatenation representation,
and report the results in \tablename{~\ref{table:SKLrepresent}}.

The comparison results show that
by using the hierarchy of dilated tree convolutions to learn the multi-level structured representation for the skeleton data in each frame,
the action prediction performance of our SSNet is significantly improved.
This clearly demonstrates the effectiveness of our newly proposed method in learning a discriminative representation of the human skeleton data in the spatial domain.

It is worth noting that,
even if we do not use the powerful multi-level structured representation, but directly use the coordinate concatenation representation,
our SSNet still outperforms the state-of-the-art skeleton-based activity analysis methods,
JCR-RNN \cite{li2016online}, ST-LSTM \cite{liu2017PAMI}, and Attention Net \cite{liu2017global}, on all the four datasets.

\begin{table*}[tbp]
\caption{Action prediction accuracies (\%) of SSNet with different skeleton representations.}
\label{table:SKLrepresent}
\centering
\small
\begin{tabular}{ccccccccccccc}
\toprule
\multirow{3}{*}{Skeleton representations} & \multicolumn{3}{c}{OAD}                & \multicolumn{3}{c}{ChaLearn Gesture}    & \multicolumn{3}{c}{PKUMMD}             & \multicolumn{3}{c}{G3D} \\
\cmidrule(r){2-13}
                                          & \multicolumn{3}{c}{Observation Ratio}  & \multicolumn{3}{c}{Observation Ratio}   & \multicolumn{3}{c}{Observation Ratio}  & \multicolumn{3}{c}{Observation Ratio} \\
\cmidrule(r){2-13}

                                          &    10\% &    50\%    & 90\%            &    10\% &    50\%    & 90\%             &    10\% &    50\%    & 90\%             &    10\% &    50\%    & 90\%      \\
\midrule
Coordinate concatenation                  &  65.6   &    79.2    & 81.6            &  17.5   &   53.5     & 65.9             &   30.0  &    68.5    & 78.6             &  70.1   & 79.1       & 82.0      \\
Multi-level structured representation &\textbf{65.8}&\textbf{81.3}&\textbf{82.8} &\textbf{19.5}&\textbf{56.2}&\textbf{69.1} &\textbf{33.9}&\textbf{74.1}&\textbf{82.9} &\textbf{72.0}&\textbf{81.2}&\textbf{83.7} \\
\bottomrule
\end{tabular}
\end{table*}

\subsection{Evaluation of Distance Regression}
\label{sec:experiment:distRegression}

We adopt the metric $SL$-$Score$ proposed in \cite{li2016online} to evaluate the distance regression performance of our network,
which is calculated as $e^{-|\hat{s} - s|/d}$,
where $s$ and $\hat{s}$ are respectively the ground truth distance and regressed distance to the action's start point,
and $d$ is the length of the action instance.
For false classification samples, the score is set to 0.


We report the regression performance of our SSNet in \tablename{~\ref{table:resultSLScore}}.
As the action detection method, JCR-RNN \cite{li2016online}, also estimates the start point, we also compare our method with it.
Besides, we investigate the regression performance of the SSNet when we do not use multi-level structured representation but directly use coordinate concatenation for it
(here we denote this case as SSNet$^C$).

The results show that our SSNet provides the best regression performance.
Specifically, we observe that the regression result of SSNet (with multi-level structured representation) is better than SSNet$^C$ (with coordinate concatenation).
This indicates that by effectively learn the spatial tree structure of the input skeleton data,
the accuracy of temporal distance regression can also be improved.


\begin{table}[tbp]
	\caption{Start point regression performance ($SL$-$Score$).
SSNet$^C$ indicates that we use the coordinate concatenation representation for the network.}
	\label{table:resultSLScore}
	\centering
	\small
	\begin{tabular}{cccc}
		\toprule
		Dataset              & JCR-RNN  &  SSNet$^C$   &      SSNet    \\
		\midrule
		OAD                  &  0.42    &   0.69    &   \textbf{0.71}     \\
        ChaLearn Gesture     &  0.49    &   0.58    &   \textbf{0.60}     \\
        PKUMMD               &  0.61    &   0.72    &   \textbf{0.75}      \\
		G3D                  &  0.62    &   0.72    &   \textbf{0.74}         \\
		\bottomrule
	\end{tabular}
\end{table}

\begin{figure}[tbp]
\begin{minipage}[b]{0.5\linewidth}
		\centering
		\centerline{\includegraphics[scale=0.216]{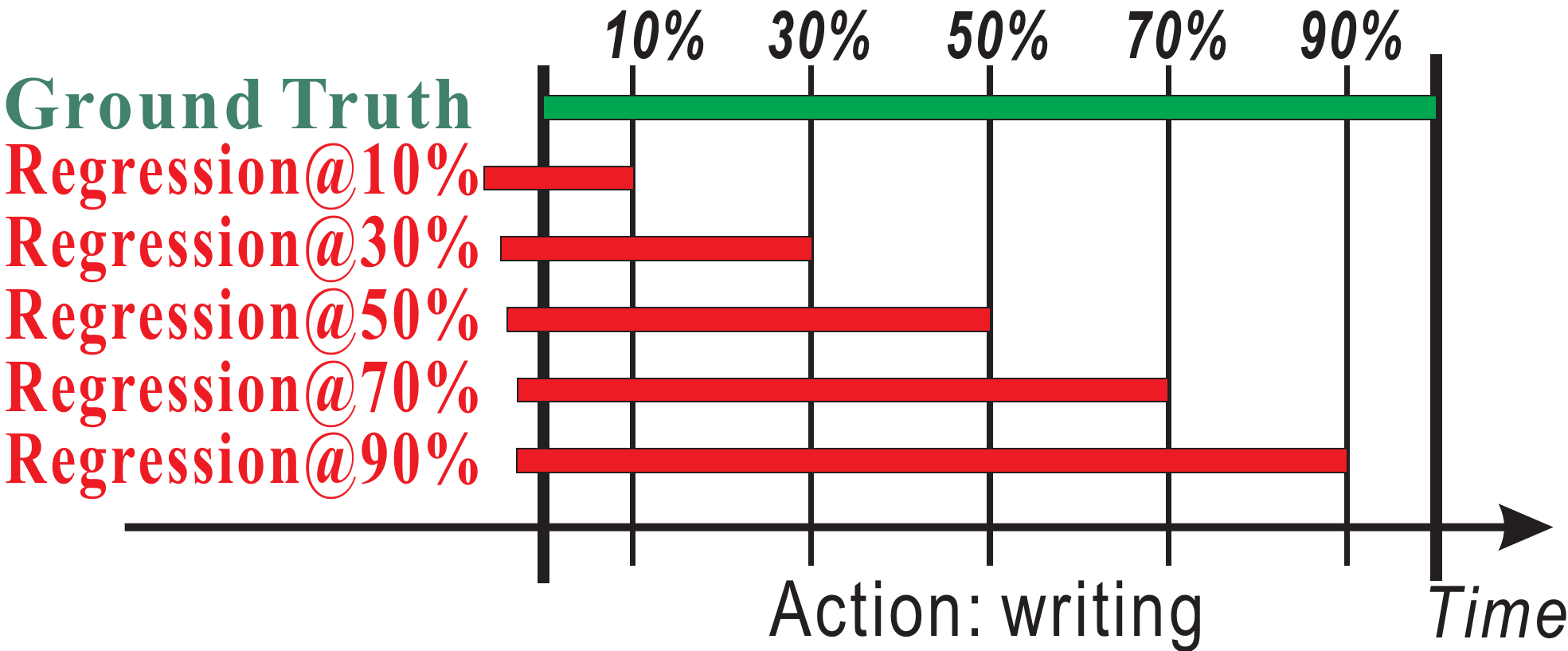}}
        (a)
\end{minipage}
\begin{minipage}[b]{0.49\linewidth}
		\centering
		\centerline{\includegraphics[scale=0.216]{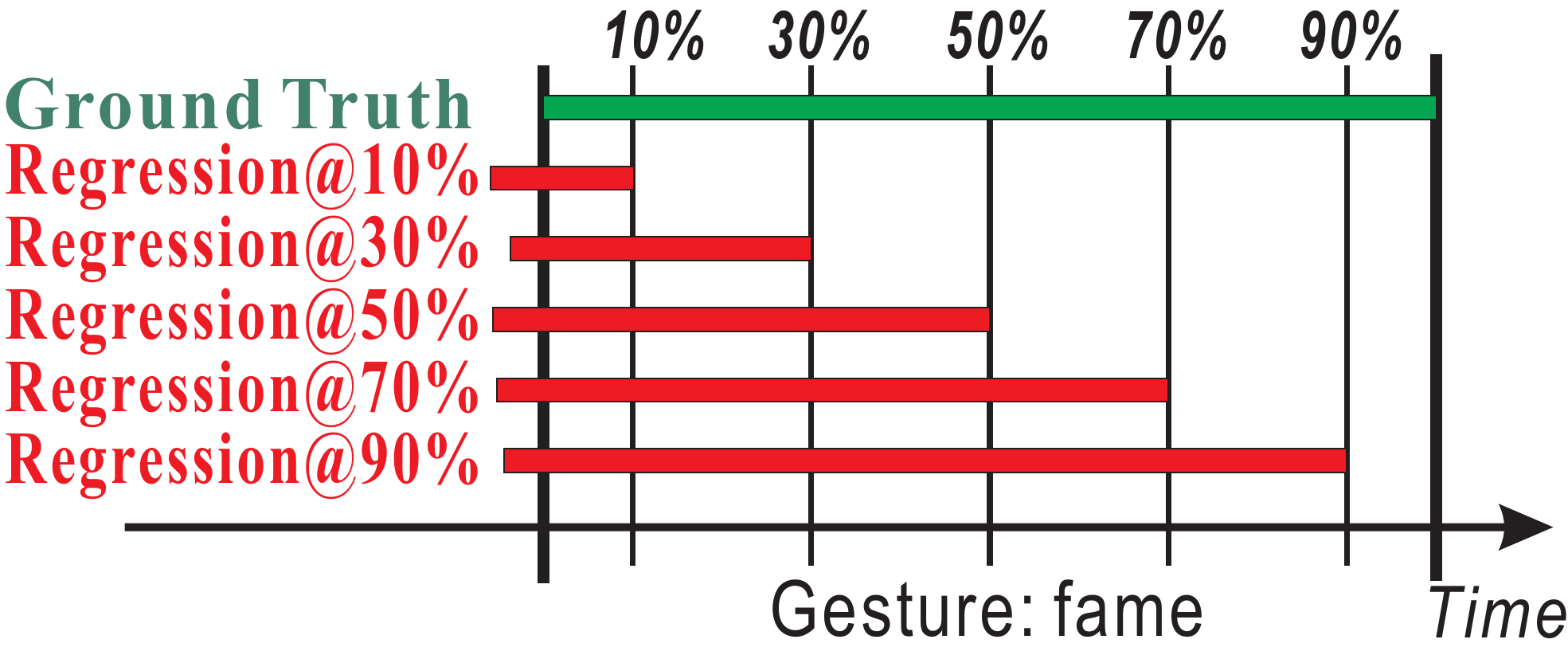}}
        (b)
\end{minipage}
\\
\\
\begin{minipage}[b]{0.5\linewidth}
		\centering
		\centerline{\includegraphics[scale=0.216]{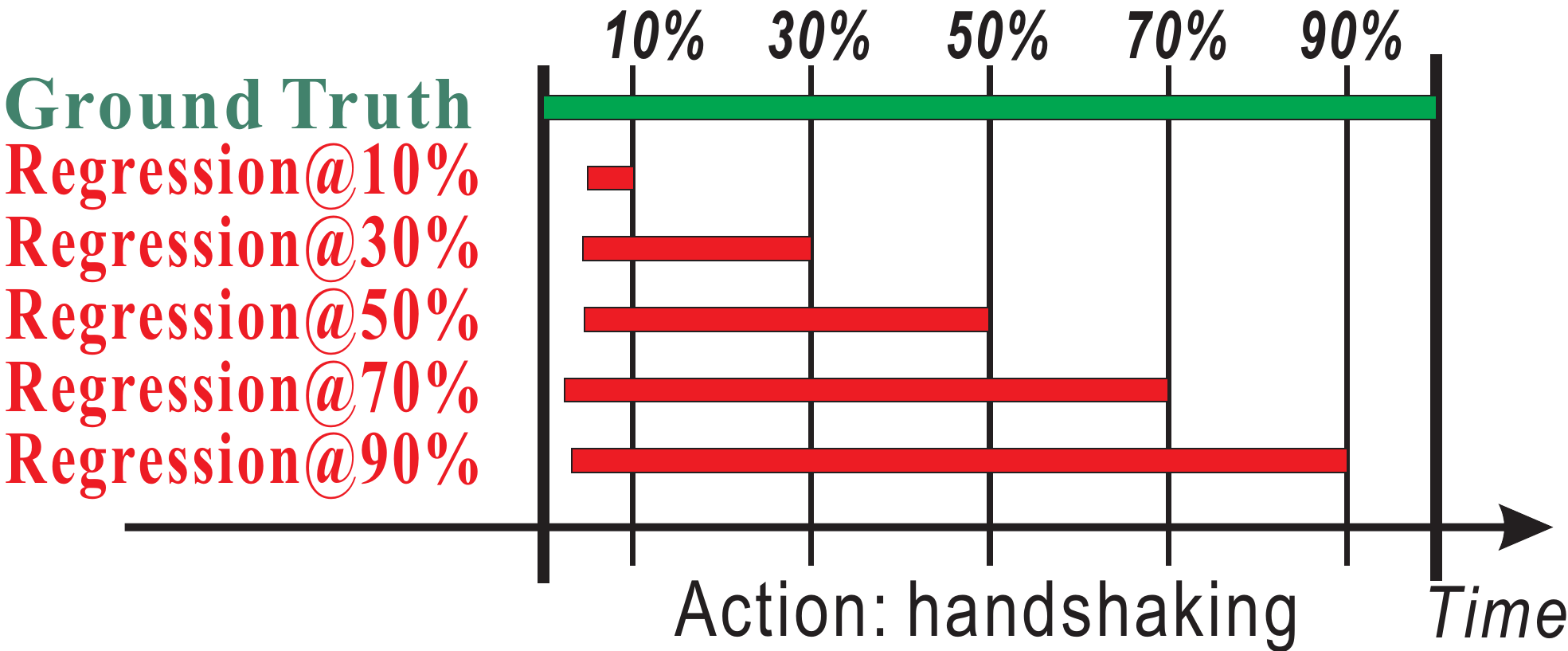}}
        (c)
	\end{minipage}
\begin{minipage}[b]{0.49\linewidth}
		\centering
		\centerline{\includegraphics[scale=0.216]{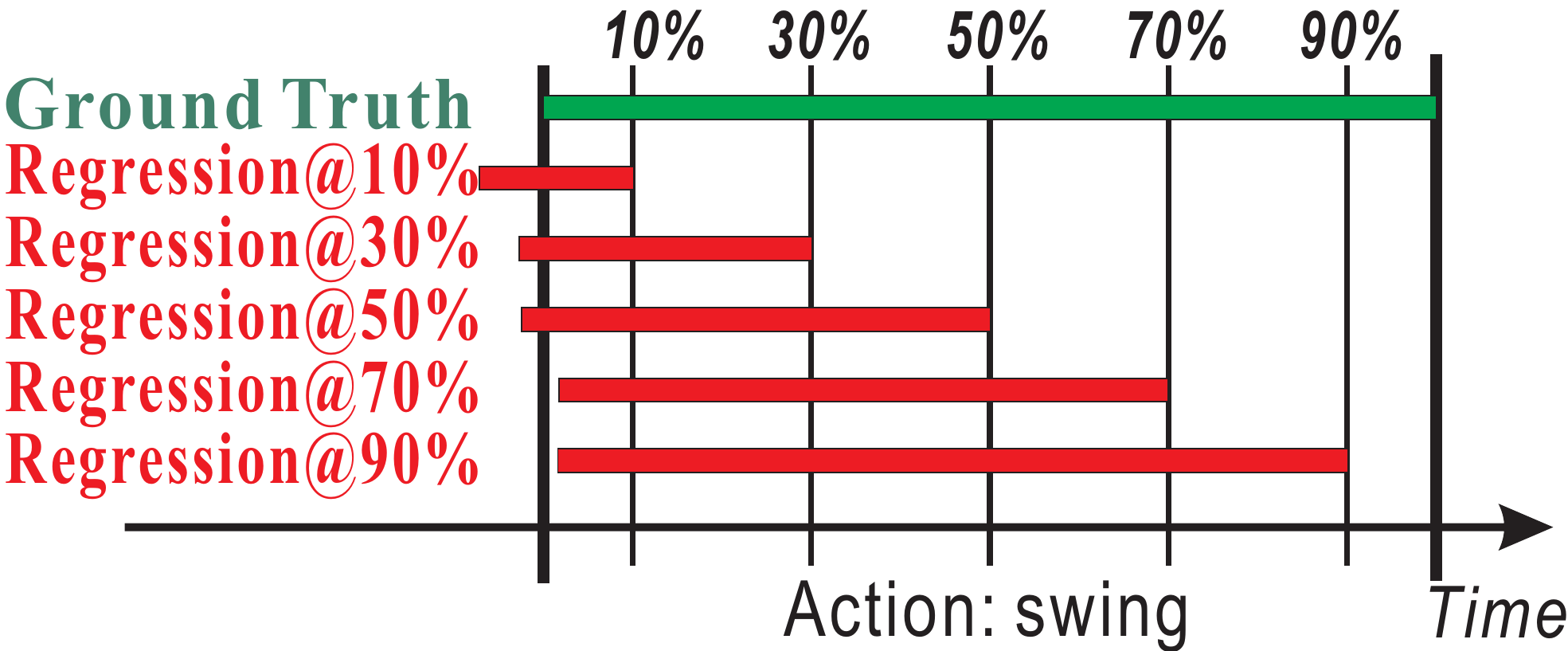}}
        (d)
\end{minipage}
	\caption{Examples of the start point regression results on the four datasets.
The leftmost point of the green bar is the ground truth start point position of the current ongoing action.
The leftmost point of each red bar (ending at $p$\%) is the regressed start point position when $p$\% of the action instance is observed.
}
	\label{fig:visualizeresult_pami}
\end{figure}

We also evaluate the average regression errors in the observed segment ($p\%$) on the large-scale ChaLearn Gesture dataset in \tablename{~\ref{table:resultRegErr}}. 
The regression error is calculated as $|\hat{s} - s|$.
We find our method regresses the distance reliably.
When only a small ratio (5\%) of the action instance has been observed, the average regression error is 6 frames.  
The regression becomes more reliable when more frames are observed.
We also visualize some examples in \figurename{~\ref{fig:visualizeresult_pami}}.
It shows that our SSNet achieves promising regression performance.


\begin{table}[tbp]
	\caption{Start point regression errors.}
	\label{table:resultRegErr}
	\centering
	\small
	\begin{tabular}{ccccccc}
		\toprule
		Observed Segment       & 5\%   &   10\%   &    30\%   &    50\%    & 70\%  &  90\% \\
		\midrule
		Error (frames)         &   6   &    4     &    3      &    3       &  3    &  3   \\
		\bottomrule
	\end{tabular}
\end{table}

\subsection{Evaluation of Network Configurations}
\label{sec:experiment:netConfig}

We configure the maximum dilation rate and the layer number to generate a set of SSNets,
which have different maximum perception window scales at the top layers.

The results in \tablename{~\ref{table:resultLayerNum}} show that
using more layers are beneficial for performance as the perception temporal window scale of the top layer increases.
However, the performance of 16 layers is almost the same as 14 layers.
A possible explanation is that the duration time of most actions is less than 255 frames.
Besides, 255 frames are long enough for activity analysis.
Thus using the SSNet with 14 layers (with maximum window scale 255) is suitable.
\begin{table*}[tbp]
	\caption{Evaluation of different configurations of the proposed network on the OAD dataset.}
	\label{table:resultLayerNum}
	\centering
	\small
	\begin{tabular}{cccccc}
		\toprule
		Number of 1-D convolutional layers                &~~~~~8~~~~~~&  ~~~~~~10~~~~~~ & ~~~~~~12~~~~~~   & ~~~~~~14~~~~~~   & ~~~~~~16~~~~~~   \\
		Maximum dilation rate                             &      8     &       16        &      32          &      64          &      128     \\
		Maximum perception temporal window scale (frames) & \emph{31}  &  \emph{63}      &    \emph{127}    &   \emph{255}     & \emph{511}   \\
		\midrule
		Start point regression ($SL$-$Score$)             &   0.65     &     0.68        &     0.70         &     0.71         &    0.71  \\
		Prediction accuracy (\%)                          &   75.2     &     77.8        &     79.0         &     80.6         &    80.6 \\
		\bottomrule
	\end{tabular}
\end{table*}

\begin{figure}[tbp]
		\centering
		\centerline{\includegraphics[trim=290 288 290 288,scale=0.520]{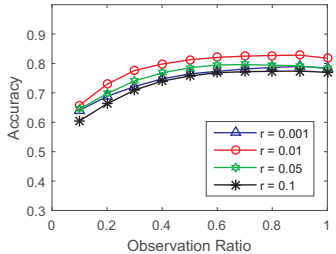}}
	\caption{Action prediction results with different $\gamma$ values on the OAD dataset.}
	\label{fig:result_curves_ChaLearn_gamma}
\end{figure}

We also evaluate the performance of our SSNet with different $\gamma$ values (see Eq (\ref{eq:objfunc}))
in \figurename{~\ref{fig:result_curves_ChaLearn_gamma}}.
We observe our SSNet yields the best performance when $\gamma$ is set to 0.01.

As shown in Eq (\ref{eq:skipconnectionC}) and Eq (\ref{eq:skipconnectionS}),
in the modules of generating representations for class prediction and distance regression,
instead of using the activation from one convolutional layer only,
we add skip connections (links from the bottom convolutional layers).
In our experiment, we observe that using this skip connection design, the action prediction accuracy can be improved by about 1.5\%.
We also investigate to further add batch normalization (BN) layers \cite{ioffe2015batch} to our network,
and we do not see obvious performance improvement,
thus BN layers are not used in our model.



\subsection{Frame-level Classification Accuracies}
\label{sec:experiment:framelevel}

As the action classification is performed at each frame of the videos,
the average classification accuracies over all frames are also evaluated, and the results are reported in \tablename{~\ref{table:resultAvgacc}}.
The results show the superiority of our SSNet over the compared approaches.

\begin{table}[tbp]
	\caption{Frame-level classification accuracies. FSNet(best) denotes the FSNet that gives the best results among all FSNets.}
	\label{table:resultAvgacc}
	\scriptsize
	\begin{tabular}{ccccccc}
		\toprule
		Dataset          &  ST-LSTM      & AttentionNet     & JCR-RNN     &  FSNet(best)  & SSNet        \\
		\midrule
		OAD              &   0.77        &    0.75          &  0.79       &    0.80       &   \textbf{0.82}      \\
        ChaLearn         &   0.62        &    0.63          &  0.62       &    0.64       &   \textbf{0.66}       \\
        PKUMMD           &   0.78        &   0.80           &  0.79       &    0.82       &   \textbf{0.85}     \\
		G3D              &   0.70        &   0.71           &  0.74       &    0.75       &   \textbf{0.76}    \\
		\bottomrule
	\end{tabular}
\end{table}


\section{Conclusion}
\label{sec:conclusion}
In this paper,
we have proposed a network model, SSNet, for online action prediction in untrimmed skeleton sequences.
A stack of convolutional layers are introduced to model the dynamics and dependencies in temporal dimension.
A scale selection scheme is also proposed for SSNet,
with which our network can choose the proper layer corresponding to the most proper window scale for action prediction at each time step.
Besides, a hierarchy of dilated tree convolutions are designed to learn the multi-level structured representations for the skeleton data in order to improve the performance of our network.
Our proposed method yields superior performance on all the evaluated benchmark datasets.
In this paper, the SSNet is proposed for handling the online action prediction problem.
This network could also be extended to address the problem of temporal action detection in streaming skeleton sequences,
which requires to locate each action in the skeleton sequence and meanwhile predict the class of each action.
We leave this extension as our future work.

\section*{Acknowledgment}

This research was carried out at Rapid-Rich Object Search (ROSE) Lab at Nanyang Technological University.
ROSE Lab is supported by the National Research Foundation, Singapore,
and the Infocomm Media Development Authority, Singapore.
This work was supported in part by the National Basic Research Program of China under Grant 2015CB351806, and the National Natural Science Foundation of China under Grant 61661146005 and Grant U1611461.
%
We acknowledge the NVIDIA AI Technology Centre (NVAITC) for the GPU donation.

\end{document}